\documentclass[a4paper]{article}

\usepackage{natbib}
\usepackage[english]{babel}
\usepackage[utf8x]{inputenc}
\usepackage[T1]{fontenc}
\usepackage{amssymb}
\usepackage{enumitem}

\usepackage[a4paper,top=3cm,bottom=2cm,left=3cm,right=3cm,marginparwidth=1.75cm]{geometry}

\usepackage{amsmath}
\usepackage{graphicx}
\usepackage[colorinlistoftodos]{todonotes}
\usepackage[colorlinks=true, allcolors=blue]{hyperref}
\usepackage{bm}
\usepackage{float}
\usepackage{rotating}
\usepackage{subcaption}
\usepackage[toc]{appendix}
\usepackage[%
   figurewithin=section,
   tablewithin=section
]{caption}

\providecommand{\keywords}[1]
{
  \small	
  \textbf{\textit{Keywords---}} #1
}

\newcommand{\X}{{\bf X}}

\newcommand{\bS}{{\bf S}}

\newcommand{\z}{{\bf z}}
\newcommand{\A}{{\bf A}}

\title{Effective degrees of freedom for surface finish defect detection and classification}
\author{Natalya Pya Arnqvist$^{1}$, Blaise Ngendangenzwa$^{1}$, Eric Lindahl$^{2}$, Leif Nilsson$^{1}$, Jun Yu$^{1}$ \\
          \small 1. Department of Mathematics and Mathematical Statistics,  Ume\aa{}  University, 901 87 Ume\aa{}, Sweden\\[0.1cm]
          \small 2. Volvo Group Truck Operations (GTO), 904 34 Ume\aa{}, Sweden\\[0.1cm]
          }

\begin{document}
\maketitle

\begin{abstract}
  One of the primary concerns of product quality control in the automotive industry is an automated detection of defects of small sizes on specular car body surfaces. A new statistical learning approach is presented for surface finish defect detection based on spline smoothing method for feature extraction and $k$-nearest neighbour probabilistic classifier. Since the surfaces are specular, %  Rather than analyzing the natural images of the car body surfaces,
  structured lightning reflection technique is applied for image acquisition. Reduced rank cubic regression splines are used to smooth the pixel values while the effective degrees of freedom of the obtained smooths serve as components of the feature vector.  A key advantage of the approach is that it allows reaching near zero misclassification error rate when applying standard learning classifiers. We also propose probability based performance evaluation metrics as alternatives to the conventional metrics. The usage of those provides the means for uncertainty estimation of the predictive performance of a classifier. Experimental classification results on the images obtained from the pilot system located at Volvo GTO Cab plant in Umeå, Sweden, show that the proposed approach is much more efficient than the compared methods. 
\end{abstract}

%\begin{keyword}
%classification \sep defect detection \sep smoothing \sep EDF \sep probabilistic $k$-NN classifier

%\end{keyword}

\section{Introduction}
Advances in the production technologies have had a great impact on automation of the majority of the production lines in the automotive industry. However, the product quality control process with only a few exceptions, yet remains a manual practice of the car body surface inspection and detection of defects. Such manual evaluation could be performed at several production stages of the manufacturing process, where specially assigned workers inspect, for example, raw surface, painted surface or surface with final finishing. Limitations and issues that occur at each production stage in connection with human inspection are very well recognized \citep{molina2017}. These include subjective human defect detection, inconsistent and  expert dependent evaluation criteria, difficulties encountered from inspecting highly reflected painted surface. The last issue is especially crucial for defects of small sizes, which are only visible when using directional light or viewed at a certain angle. %Therefore, to replace labour intensive manual inspection and at the same time to provide with persistent reliable evaluation that will lead to increase in production efficiency and product quality, by means of a systematic approach to automated defect detection is the most important and challenging task for automotive industry general managers.  
Therefore, replacing labor intensive manual inspection and providing with persistent reliable evaluation is one of the most important and challenging tasks for automotive industry general managers. A systematic approach to automated defect detection will lead to increase in production efficiency and product quality, and thereby lowering labour cost, reducing the need of repair and adjustments, and reducing environmental impact.
%\todo[inline,color=yellow]{below is unfinished literature review; the idea is to write first lit review on stats approaches, then mention machine vision techniques/Natalya} 

Several systems have been developed during the last decade to provide a solution to this problem. Deflectometry-based detection on specular surfaces has proven to be a reliable and accurate approach to accomplish the task of detecting defects on car body surfaces \citep{fraunhofer, kammel2008}.  Approaches using deflectometry- and vision-based technologies combined with image fusion \citep{fotsing2014,leon2006,stathaki2011} have been installed in automotive industry, for instance Ford \citep{armesto2011}, Opel \citep{santolaria2016} and Mercedes-Benz \citep{micro-epsilon}. Deep statistical learning and advanced statistical modelling of high-dimensional spatio-temporal data, combined with machine colour-based vision and image analysis, are very useful for image segmentation and pattern recognition \citep{bishop2007,severino2013}, e.g. recognizing defects on painted vehicle bodies \citep{maestro2018,sikandar2018,pya2018}. An analysis based on only one image channel (sensor) is often insufficient and therefore combining data from multiple sensors or image channels is crucial to obtain the desired information \citep{heizmann2011,weckenmann2009,yu2003}.

Here we develop a statistical learning approach for defect detection on painted cab surfaces, covering image acquisition, feature extraction and defect classification. The aim is to develop an approach that not only produces %most 
accurate and reliable classification but also provides uncertainty assessment for the classification results. To accomplish this we build feature descriptors using regression spline ideas applied to grey intensity pixel values of the acquired images. As the inspection of specular surfaces inflicts special challenges,  deflectometry %structured lightning
technique using reflected sinusoidal fringe is applied to capture images of the considered surfaces. Classification is then achieved using a probabilistic classification algorithm based on $k$-nearest neighbour classifier \citep{ranneby2011nonparametric}.

This paper contributes the following novel elements necessary to succeed in defect detection and classification.
\begin{enumerate}
    \item We propose novel feature descriptors based solely on the smoothness degree of the fitted splines. These allow avoiding the usage of classification algorithms with computational expensive training phases that are most commonly used for surface defect detection. Instead we apply the probabilistic classifier based on $k$-nearest neighbour algorithm that results in highly accurate and reliable classification.
    \item We provide a nonparametric patchwise probabilistic classification approach built upon the one nearest neighbour rule with Euclidean distance. The estimates of the proper probabilities for each class are obtained using the concept of NN-balls in the feature space. 
    \item The probability based performance evaluation metrics are presented as alternatives to such conventional metrics as misclassification error rates, false positive and false negative rates. Moreover, the vectors of posterior probabilities of the considered classifier allow for classification quality assessment in terms of the uncertainty measure, Entropy. 
\end{enumerate}

The remainder of the paper is organized as follows. In Section \ref{section:approach} the proposed statistical learning approach for defect detection is described. We outline the experimental industrial setup to which our approach was applied, discuss the performance results and present the comparative study in Section \ref{section:results}. We give our conclusions in Section \ref{section:conclusions}.

\section{Defect detection approach}\label{section:approach}
The illustration of the proposed detection procedure is split into three parts. Firstly we briefly explain the technique used for image acquisition. The extraction of novel features is described in subsection \ref{sec:features} which is followed by a description of the learning classifier applied.

\subsection{Deflectometry-based image acquisition}\label{section:deflectometry}
Our approach for defect detection relies on smoothing features extracted from the images acquired by the deflectometry technique. Inspection of the specular surfaces using deflectometry principle has been an effective approach in the field of machine vision. The technique uses the geometry of specular reflection to measure the gradient of the inspected surface. Typically, a machine vision approach involves four stages similar to those of the statistical learning approach, such as image acquisition, image processing, feature extraction and classification. The proposed statistical approach borrows the image acquisition part from the deflectometry-based machine vision technique. 

\begin{figure}[H]
\begin{centering}
\includegraphics[width=0.6\textwidth]{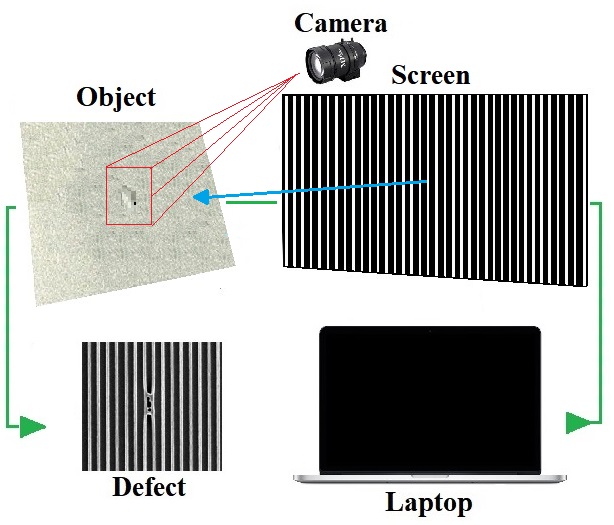}
\par\end{centering}
\caption{\label{fig:deflectometry}A sketch of the deflectometry-based image acquisition process. }
\end{figure}

The basic components of the machine vision system include a camera, a screen, illuminations, image processing hardware and software. The idea is to project structured light patterns (the most commonly used pattern is sinusoidal) from the screen over an inspected surface and observe the specular surface reflection of the patterns captured by the camera. Any changes of the surface lead to distortion of the observed reflection from the reflection that a defect-free surface would otherwise show. Figure \ref{fig:deflectometry} displays a schematic setup for image acquisition based on the deflectometry principle. Sinusoidal fringes are displayed on the screen/monitor, and the specular reflection of the patterns by the inspected surface is captured by the camera. The general applicability of deflectometry requires a well-calibrated setup which considers camera parameters and some geometric parameters that bring information between the camera, screen and the surface. Here we apply the three-step system calibration method proposed in \cite{knauer2004}.  This method divides the system calibration into camera calibration step, screen calibration and geometric calibration steps. The discussion of each step is omitted here as it is beyond the scope of this paper (see  \cite{knauer2004} for further details).  

The sinusoidal pattern displayed on the screen is obtained using the following mathematical expression.
$$I=B + A\sin(2\pi f q +\psi),$$ 
%\todo[color=yellow]{the formula needs to be checked with the one implemented in Matlab. Blaise, could you do it?/Natalya}
%\todo{done it. /Blaise}
where $B$ denotes an offset, $A$ is an amplitude, $f$ stands for a frequency, %$q$ indicates a constant 
and $\psi$ is a phase shift. %In this setting, $\psi ~\in ~\{0,\pi/2,\pi,3\pi/2\},$ $f~\in~\{8,16,32,64\}$

\begin{figure}[H]
\centering
\includegraphics[width=.9\textwidth,height=.35\textheight]{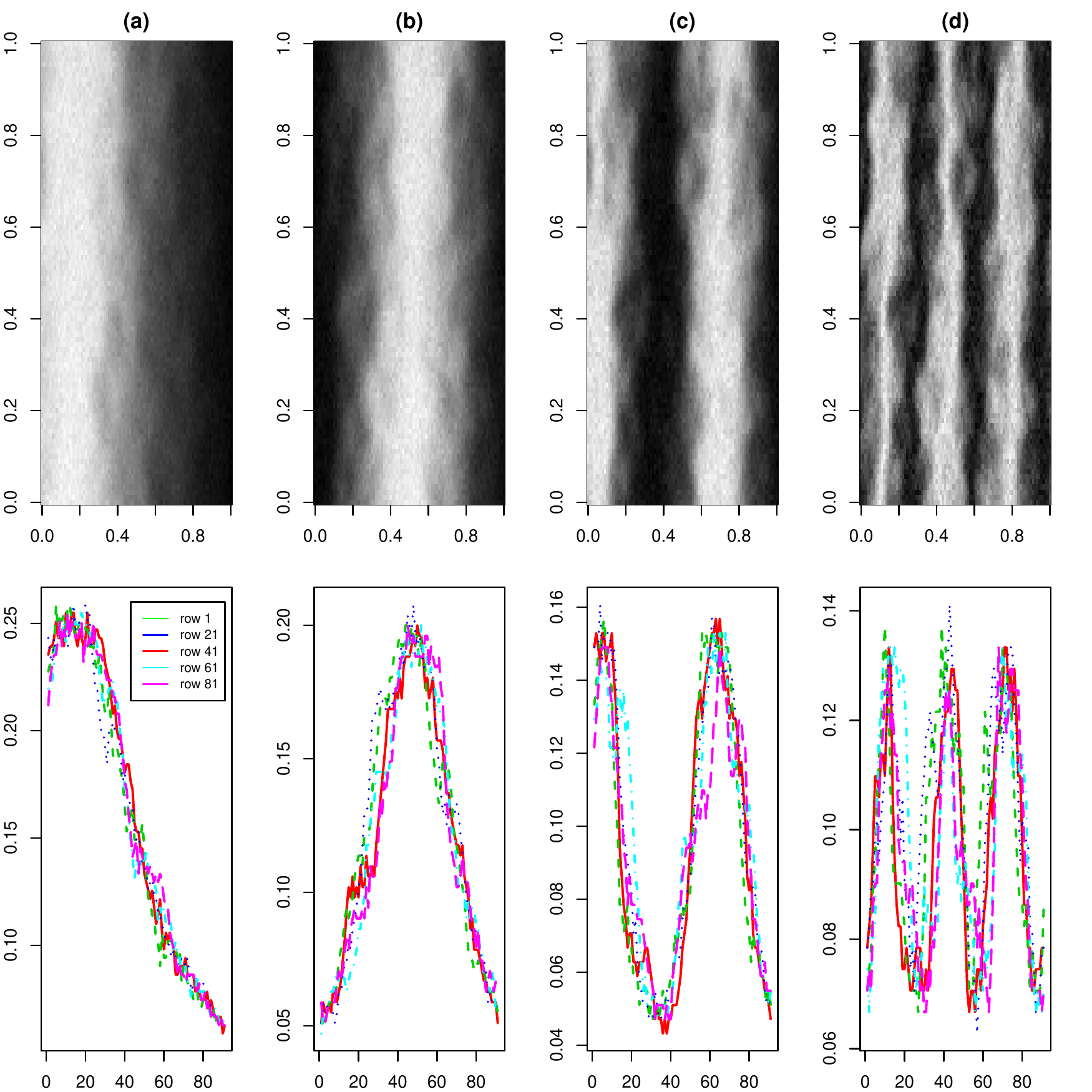}
\caption{\label{fig:nondefect-examples}Examples of the image patches of the projected pattern reflected by a defect-free cab surface.  Top row: original patches when the phase parameter of the sinusoidal pattern is set to $3\pi/2,$ but  the frequency parameter assumes four different values: (a) $f= 8$, (b) $f= 16$, (c) $f= 32$, and (d) $f= 64$. Bottom row: grey intensity pixel values of every twentieth row (rows 1, 21, 41, 61, and 81) of the corresponding patches.}
\end{figure}

\begin{figure}[H]
\centering
\includegraphics[width=.9\textwidth,height=.35\textheight]{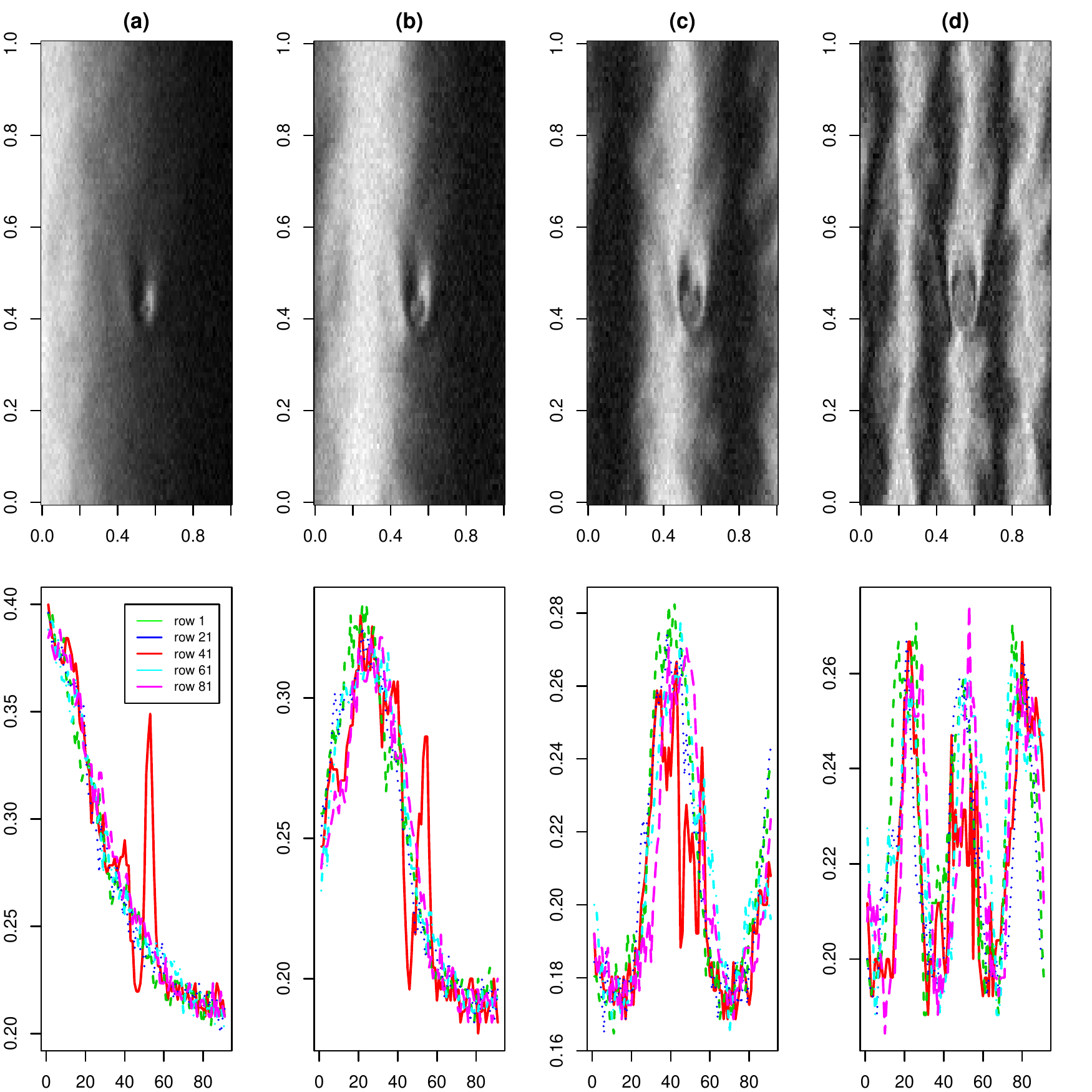}
\caption{\label{fig:crater-examples}Examples of the image patches of the projected pattern that is distorted due to cab surface defect.  The type of defect of the inspected surface is crater. Top row: original patches containing crater when $\psi=3\pi/2,$ and: (a) $f= 8$, (b) $f= 16$, (c) $f= 32$, and (d) $f= 64$. Bottom row: grey intensity pixel values of every twentieth row of the corresponding patches.}
\end{figure}

\begin{figure}[H]
\centering
\includegraphics[width=.9\textwidth,height=.35\textheight]{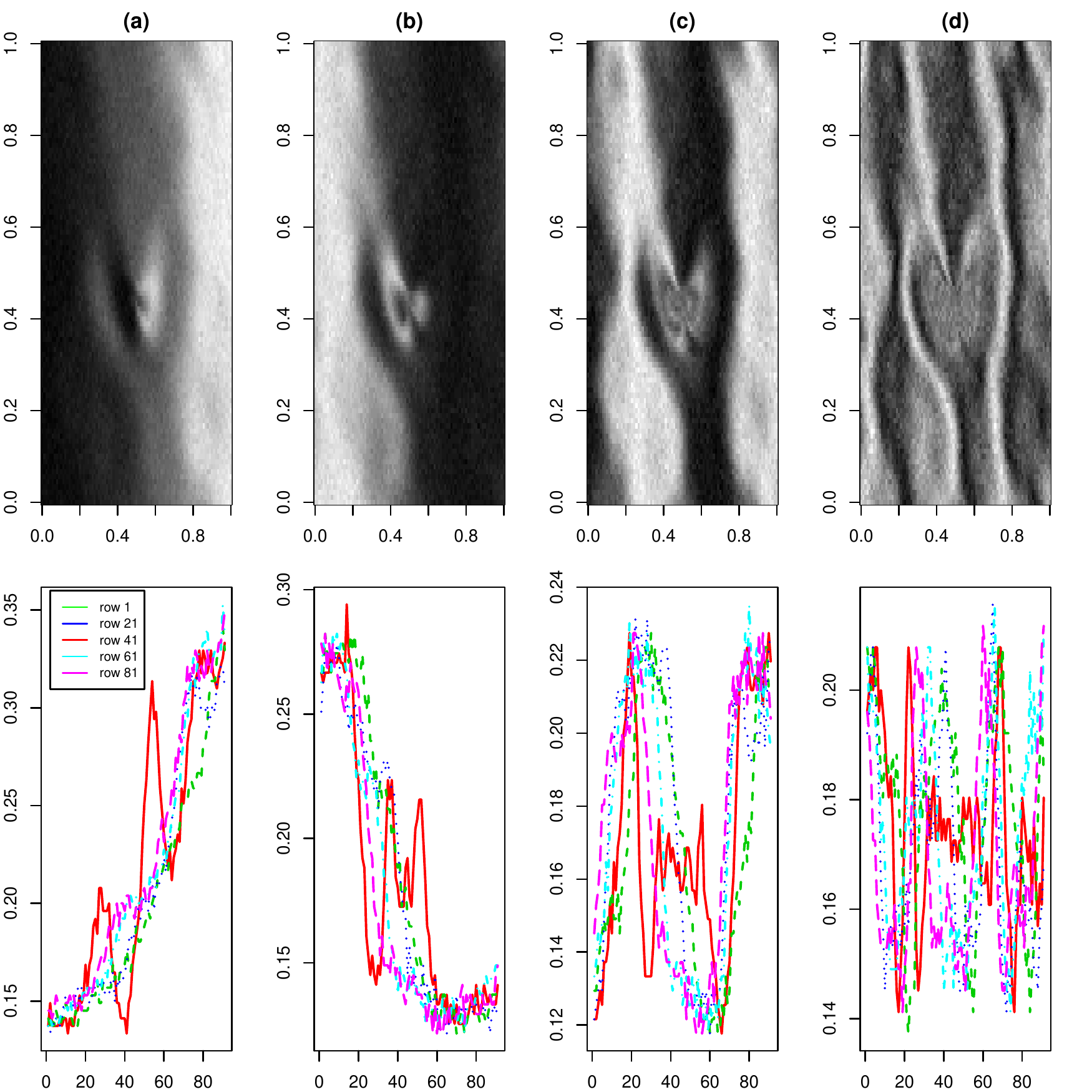}
\caption{\label{fig:dirt-examples}Examples of the image patches of the projected pattern that is distorted due to cab surface defect.  The type of defect of the inspected surface is dirt. Top row: original patches containing dirt when  $\psi=3\pi/2$ and: (a) $f= 8$, (b) $f= 16$, (c) $f= 32$, and (d) $f= 64$. Bottom row: grey intensity pixel values of every twentieth row (rows 1, 21, 41, 61, and 81) of the corresponding patches.}
\end{figure}

Rather than examining the sample images of the pattern reflection, 
we propose to use a dataset of image patches of the size $m \times m,$ extracted from the captured images. $m$ is the number of pixels in $x-$ and $y$ directions. So, the dataset will consist of image patches labeled as defect or defect-free. Ideally, one would wish to annotate defects on pixel level and then conduct classification task on that level. But it turns out in reality that it is very difficult (if not impossible) to mark each defect at pixel level, and therefore our data are collected on patch level. 
Figures \ref{fig:nondefect-examples}-\ref{fig:dirt-examples} show examples of the sample patches of size $91\times 91$ of the pattern reflection by a defect-free cab surface and by surfaces with such defects as crater and dirt,  respectively. Four image patches in the top rows of each figure correspond to captured reflections of the sinusoidal pattern with four different values of the frequency parameter, $f=8, ~16,~32, ~64,$ projected onto the same surface. 
The plots of the bottom rows illustrate grey intensity pixel values of the five selected rows (every twentieth row) for each patch. %The grey intensity pixel values of the selected rows of each images shown in the bottom plots, illustrate the variability of the intensities between the pattern reflection of the defect-free surface and distorted reflection due to surface defects.
One may note how much wigglier the intensity values of the middle row (row 41) are for the patches with defects shown as solid red lines in Figures \ref{fig:crater-examples} and \ref{fig:dirt-examples}, in comparison with the intensities of the same row for the defect-free patch shown in Figure \ref{fig:nondefect-examples}. It seems, therefore, natural to try to capture that wiggliness when building features for further classifier training. The construction of the proposed feature vectors is presented in the next subsection.
%\todo[inline]{need to clarify the figures 2-4 and figure caption as follows: i) not necessary with additional (e)-(f). it will be enough with (a)-(d) for the frequency parameter. ii) need captions for the different colored curves. /Jun}

%structured lightning reflection techniques     

\subsection{Smoothing features}\label{sec:features}

Consider a patch of size $m \times m$. The idea is to reproduce the projected sinusoidal patterns by smoothing pixel values row-wise using penalized regression splines, and to pick up the wiggliness of the smoothed patterns using the effective degrees of freedom (EDF) of the fitted splines. To do so, we build a semiparametric Gaussian model for each row of the intensity values,  
  $$
    z_{rj}=g_r(t_j) +\epsilon_j, \quad r, j =1, ..., m,
  $$
where $z_{rj}$ denotes the value of the pixel $(r, j)$ in the patch, $t_j=j,$ $g$ is an unknown smooth function, and $\epsilon_j$'s are independent $N(0,\sigma_j^2)$ random variables. There are several alternatives available when choosing univariate penalized regression spline to estimate $g_r(t).$ However, almost all of them produce similar results. Cubic regression splines are chosen here due to their sufficient flexibility and efficiency. This spline can be written in the following form \citep{wood2017},
  $$
   g_r(t)=\sum\limits_{k=1}^qb_k(t)\beta_{rk},
  $$
where $b_k(t)$ are known basis functions, $\beta_{rk}$ are unknown coefficients to be estimated, and $q$ is the number of basis functions used. When using any smoothing technique, the natural question arises of how to select the degree of model smoothness. This is controlled by $q$ in this case. Instead of choosing the  smoothness degree by selecting $q$, the standard practice is to control the model's smoothness by adding `wiggliness' penalty to the least squares fitting objective. At the same time $q$ is kept fixed at a sufficiently large size to avoid oversmoothing. Thus, to estimate $\beta_{rk}$ the following penalized regression fitting objective is minimized 
\begin{equation}\label{pls1}
   ||\z_r-\X\bm\beta_r||^2+\lambda_r \int\limits_{z_1}^{z_k}g_r''(t)^2dt,
\end{equation}
where $\z_r=(z_{r1},\ldots,z_{rm})^T,$ $\X$ is the model matrix evaluating spline basis functions at the observations, i.e. elements of $\X$ are $X_{jk}=b_k(t_j).$ $\bm\beta_r=(\beta_{r1},\ldots,\beta_{rq})^T,$ $\lambda_r$ is a smoothing parameter that controls the balance between smoothness of the fitted curve $g_r$ and data fit. It can be shown that the penalty term in (\ref{pls1}) can be represented as $\int\limits_{z_1}^{z_k}g_r''(t)^2dt=\bm\beta_r^T \bS\bm\beta_r,$ where $\bS$ is the penalty matrix of known components for the basis \citep{wood2017}, $z_1$ and $z_k$ are the end knots of the cubic spline function. 
The choice of $\lambda,$ can be made by a generalized cross validation score
$$
  \mathcal{V}_g=\frac{m\sum\limits_{j=1}^m(z_{ri} - \hat{g}_{rj})^2}{\left[m-\textrm{tr}(\A_r)\right]^2},
$$
where $\A_r=\X\left(\X^T\X +\lambda_r\bS \right)^{-1}\X^T$ is a model's influence (hat) matrix, and $\tilde{\tau}_r=\textrm{tr}(\A_r)$ is then its EDF. 
%\todo[inline]{some notation issues: $z_i$ is new, $S$ is new, $X_{kj}$ should be $X_{jk}$, the integral limits in eq (1), /Jun}

\begin{figure}
   \centering
    \includegraphics[width=.8\textwidth,height=0.3\textheight]{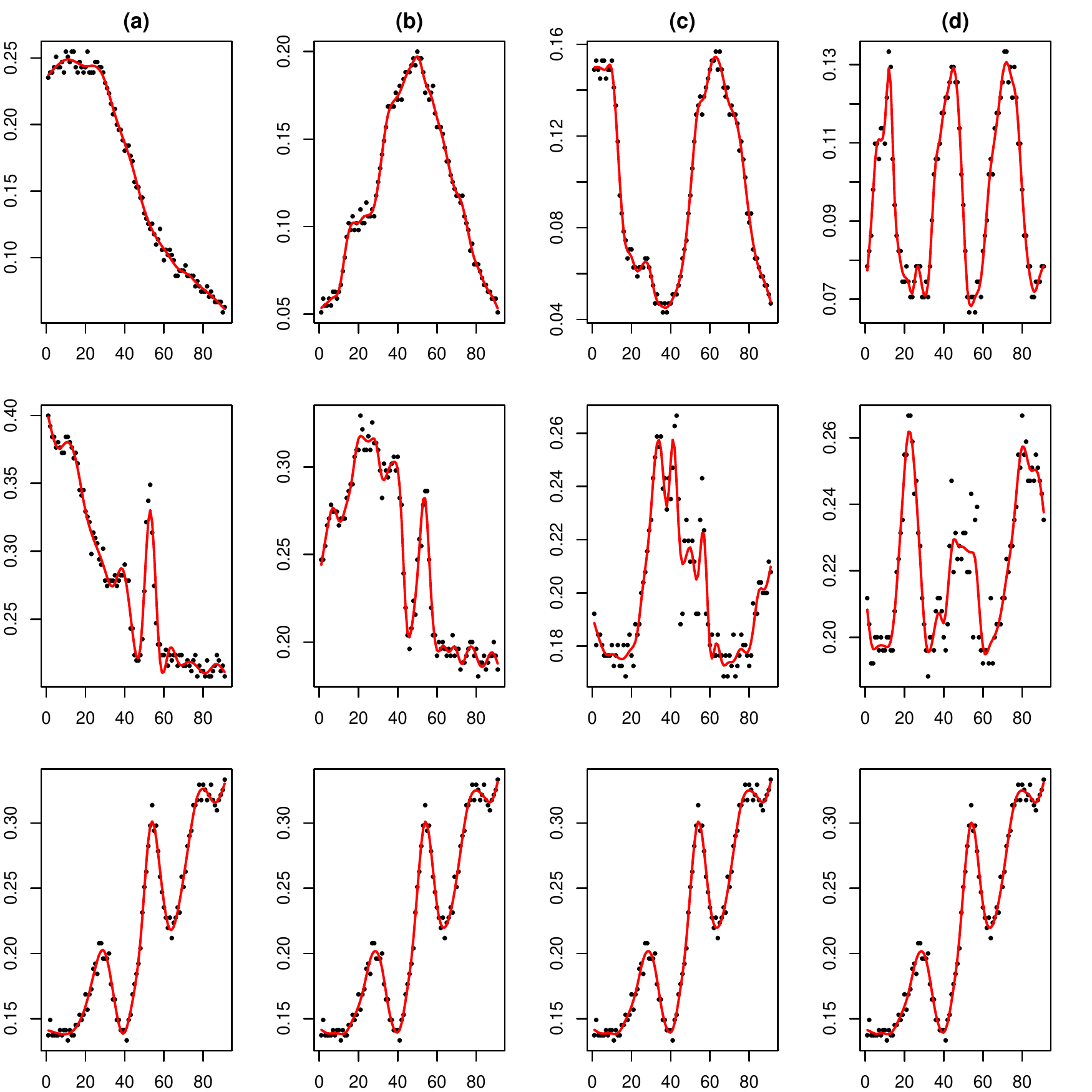}
     \caption{\label{fig:smoothed-row41}(a)-(d) Smooths of the intensity values of the $41^{\textrm{th}}$ row for the defect-free image patches, for the patches with crater, and with dirt, together with the corresponding intensity values,  for the four different values of the frequency parameter. Top row: smooths of the four defect-free patches (as in Figure \ref{fig:nondefect-examples}). Middle row: smooths of the four patches with crater (as in Figure \ref{fig:crater-examples}). Bottom row: smooths of the four patches with dirt (as in Figure \ref{fig:dirt-examples}).}
   \end{figure}

  Figure \ref{fig:smoothed-row41} shows examples of the obtained smooths of the pixel values of the $41^{\textrm{th}}$ row for the image patches in Figures \ref{fig:nondefect-examples}-\ref{fig:dirt-examples}. The degree of smoothness depends on the frequency parameter of the projected sinusoidal pattern. Based on some preliminary analysis, $q$ set to $20$ is considered to be large enough when smoothing patches with the frequency parameter, $f=8;$ $q=30$ for patches with $f=16$ and $f=32;$ and $q=40$ for the frequency $64.$
%\todo[inline]{the same for fig 5 as fig 2-4, skip (e)-(l) and use only (a)-(d) to denote the four different frequency parameter /Jun}

\begin{figure}
   \centering
    \includegraphics[width=.8\textwidth,height=0.3\textheight]{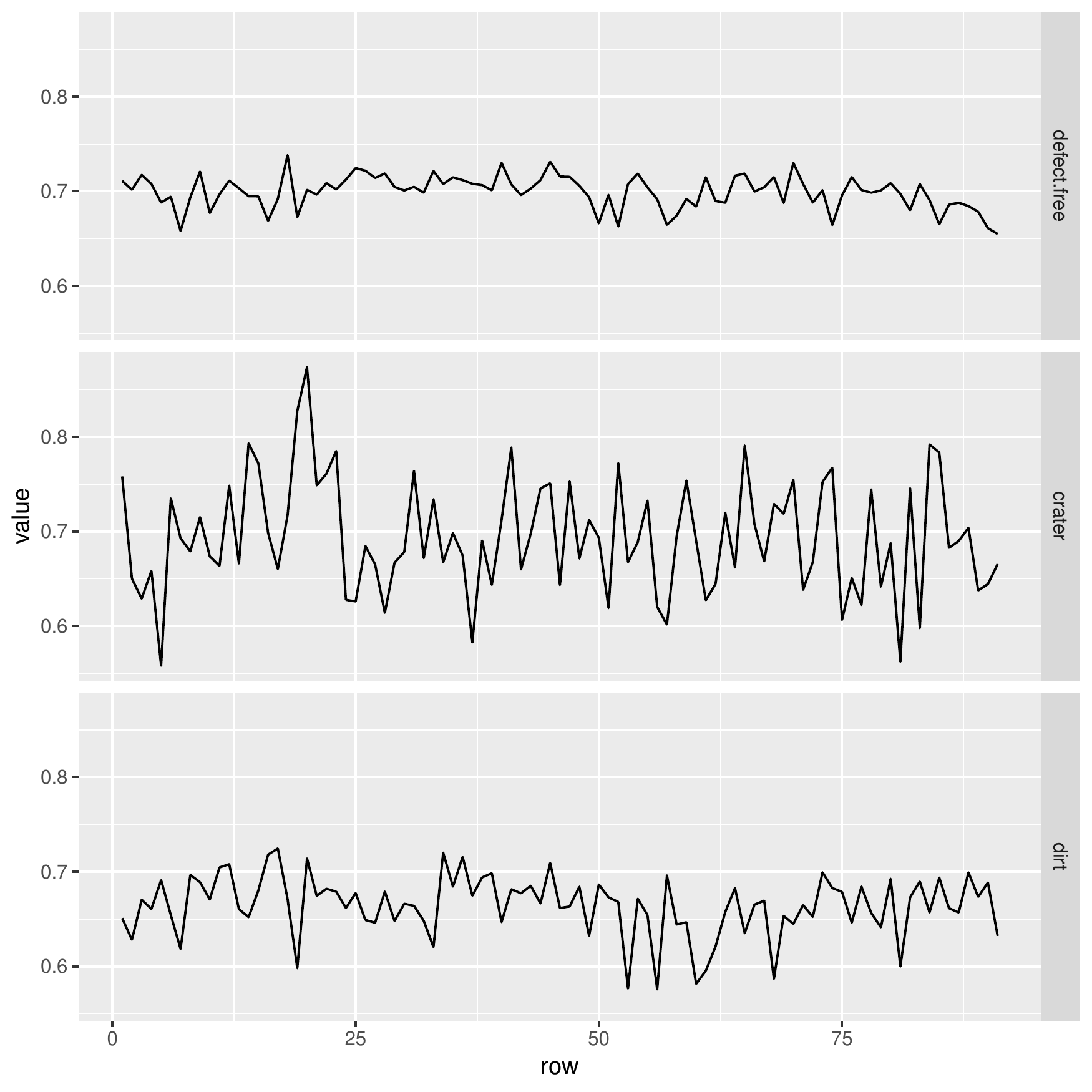}
     \caption{\label{fig:feature-vector-means}Feature vector means for each class.}
   \end{figure}

As expected, the distortion of the pattern caused by defects results in some irregularities in the obtained smooth functions. The number of the effective degrees of freedom of the model can reflect on that, with wigglier, more irregular curves resulting in larger values of $\tilde{\tau}.$ Hence, the EDF $\tilde{\tau}$  is treated as a highly plausible feature, and the combination of $m$ scaled EDFs, obtained from smoothing the image row-wise, forms our proposed feature vector,
 $$
  \bm\tau=(\tau_1,\ldots, \tau_m)^T, ~~~ \textrm{where} ~~
  \tau_r=\frac{\tilde{\tau}_r}{\max_r \tilde{\tau}_r}.
 $$
 The reason for applying the feature scaling is twofold, implying that it enables better discrimination between classes in some cases and also reduces training time of classification algorithms. Figure \ref{fig:feature-vector-means} illustrates the class feature vector means. For a fair comparison of the means of each class, three hundred images were randomly selected within each class for the means calculations. The feature vector means can be viewed as class centroids similar to those in $k$-means clustering, serving as a prototype of each class. Note that a very distinct contrast between class means supports the idea of extracting EDFs. 

As different paint colors and various finishing paint processes on the cab surface are known to affect the gray scale pixel values, the pixel values are standardized before smoothing by subtracting the overall mean of the pixel values of the patch and dividing by the overall standard deviation. 
%An R package \verb+scam+ \cite{scam18} is used for smoothing pixel values.

\subsection{Learning classifier: Probabilistic k-NN}

The constructed feature vectors $\bm\tau$ with their associated class labels can further be used to train a classification algorithm. The standard $k$-NN is one of the most straightforward nonparametric methods to classifying objects on the basis of the feature vectors which are considered as points belonging to certain class, in the feature space $\mathbb{R}^m.$ With the $k$-NN rule, class prediction is performed by finding the $k$ nearest (in some distance metric) points and assigning the most frequent label. Despite its simplicity, the performance of $k$-NN shown on numerous classification tasks signifies that it continues to be a competitive classification method in machine learning and statistics \citep{ripley2007pattern,zhang2018efficient}. However, there are a number of weak points with the $k$-NN rule, the main one being the lack of a probabilistic basis for the statistical inference which would allow, for example, employing uncertainty measures associated with the assigned class label. This issue has been addresses in several papers that introduce probabilistic alternatives of the $k$-NN method  (\cite{holmes2002probabilistic,cucala2009bayesian,friel2011classification} among others). Here we apply a probabilistic type of $k$-NN proposed by \cite{ranneby2011nonparametric} that offers a straightforward and yet theoretically neat variant of probabilistic $k$-NN classifier. The following shortly sketches the classification approach.

Consider a data sample $\left\{(c_1,\bm\tau_1),\ldots,(c_N,\bm\tau_N) \right\},$ where $c_n\in \{C_1,\ldots,C_K\}$ is a class label associated with the feature vector $\bm\tau_n\in \mathbb{R}^m.$  In our case there would be three classes: class $C_1$ of defect-free patches, class $C_2$ of patches with crater, and $C_3$ of patches with dirt. 

Let $B(\bm\tau_p,r)=\{\bm\tau: ~~|\bm\tau_p-\bm\tau|\leq r\}$ denote a ball of radius $r$ with a centre at $\bm\tau_p$ in the feature space. $\bm\tau_p$ is treated as a new test datum with an unknown class label that we would like to predict. The volume of the ball $B(\bm\tau_p,r)$ can be calculated as  
    $$V_B(r)=a_m\cdot r^m,$$
where $a_m=\pi^{m/2}/\Gamma(m/2+1),$ $m$ is the dimension of the feature vector and $\Gamma(\cdot)$ is gamma function.
Let further $D(\bm\tau_p,C_j)=\min\limits_{C_j}||\bm\tau_p-\bm\tau_{C_j}||$ denote the minimum Euclidean distance  in the feature space from $\bm\tau_p$ to the points of class $C_j.$
  
Following \cite{ranneby2011nonparametric} it can be shown that the conditional probability density for class $C_j$ is estimated by 
$$\hat{p}_j(\bm\tau_p)=\frac{1}{n_j\cdot V_B(D(\bm\tau_p,C_j))}.$$
This affords the following posterior class probabilities that allows us to assign proper probabilities to all classes:
$$p^j_p= \hat{p}(C_j|\bm\tau_p)=\frac{D^{-m}(\bm\tau_p,C_j)}{\sum_{i=1}^K D^{-m}(\bm\tau_p,C_i)}, \quad j=1, ..., K.$$
Therefore, the class belonging probability vector for a new point with feature vector $\bm\tau_p$ can be written as 
%$${\bf p}(\bm\tau_p)=\left[p(C_1|\bm\tau_p),\ldots, p(C_s|\bm\tau_p) \right].$$
$${\bf p}(\bm\tau_p)=\left[p^1_p,\ldots, p^K_p\right].$$

\subsection{Performance evaluation}

The performance of a classifier can be evaluated with various metrics where preference for particular metrics can be problem-specific. Here we use probability based performance evaluation metrics that are regarded as alternatives to such conventional metrics as misclassification error rate, false positive and false negative rates.  The estimates of the posterior probabilities of class membership, ${\bf p}(\bm\tau),$ that are supplied as the classifier outputs together with class labels, form the basis for the proposed evaluation criteria. Consider for simplicity a binary classification problem given a data sample $\left\{(c_1,\bm\tau_1),\ldots,(c_N,\bm\tau_N) \right\},$ where $c_n\in \{C_0,C_1\},$ $C_0$ denotes the class of defect-free patches and $C_1$ class of defects. Let $N_1$ denote the number of patches with defect and $N_0$ number of patches without defects. Denote the posterior probability, the conditional probability that the patch has a defect (or it is defect-free) given the feature vector $\bm\tau_n$ by $p^1_n=\textrm{P}(C_1|\bm\tau_n)$ ($p^0_n=\textrm{P}(C_0|\bm\tau_n)$ for defect-free). For the binary classification $p^0_n+p^1_n=1.$ Then the probability of misclassification, PM$_n$, can be calculated as PM$_n=1-p^1_n$ if the patch really has a defect (belongs to class $C_1$) and PM$_n=p^1_n$ if the patch is defect-free (belongs to class $C_0$). Furthermore the probability of false negative is defined as PFN$_n=1-p^1_n$ for the patches with defects, and the probability of false positive  as PFP$_n=p^1_n$ for defect-free patches. Note that PFN for defect-free patches and PFP for defects are not defined. The introduced notations are summarized in the following Table \ref{tab:prob-metrics}. 

 \begin{table}[H]
\centering \caption{Error probabilities\label{tab:prob-metrics}}
\setlength\arrayrulewidth{1pt}
\begin{tabular}{cccccc}
\hline
 $n$ & Label & $p^1_n=\textrm{P}(C_1|\bm\tau_n)$ & $\textrm{PM}_n$ & 
 $\textrm{PFN}_n$ & $\textrm{PFP}_n$\\
\hline
  1  &$C_1$  & $p^1_1$  & $1-p^1_1$     &  $1-p^1_1$  &  --      \\
  2  &$C_0$  & $p^1_2$  & $p^1_2$       &  --       &  $p^1_2$    \\
 % 3  &$C_1$  & $p^1_3$  & $1-p^1_3$     & $1-p^1_3$   &  --       \\
  .  &.  &  .       & .         &  .        &    .      \\
  .  &.  &   .      & .         & .         &    .      \\
% $n$ &0  &$p_n$   & $p_n$       & --   &  $p_n$        \\
\hline
\end{tabular}
\end{table}
The performance criteria similar to the conventional metrics but based on the probabilities can now be specified as follows.
\begin{enumerate}[label=\alph*)]
\item Probability based misclassification error rate: 
$\textrm{probMER}= \frac{1}{N}\sum\limits_{n=1}^N\textrm{PM}_n.$
\item  Probability based false negative rate:
$
 \textrm{probFNR}= \frac{1}{N_1}\sum\limits_{n=1}^{N_1}\textrm{PFN}_n.
$
\item  Probability based false positive rate:
$
 \textrm{probFPR}= \frac{1}{N_0}\sum\limits_{n=1}^{N_0}\textrm{PFP}_n.
$
\end{enumerate}

Moreover, the vectors of posterior probabilities allow the classification quality to be judged at patch level, by using the uncertainty measure via Shannon entropy \citep{ranneby2011nonparametric}:
$$ 
  H(\bm\tau_n)=-\left(p^0_n\cdot \log p^0_n + p^1_n\cdot \log p^1_n\right).
 %  H(p_{c_j})=-\sum\limits_{j=1}^sp_{c_j}\cdot \log p_{c_j}
$$
Therefore, the uncertainty of  classification for the whole dataset can be measured by the average entropy:
$$ 
  H=-\frac{1}{N}\sum\limits_{i=1}^N H(\bm\tau_n).
$$
The above mentioned performance metrics can be easily extended to multi-class classification problems. Probabilities of false positive and false negatives would be obtained by applying one-against-all principle that evaluates these probabilities relatively to one class. However, for the defect classification problem discussed in this paper, probability based false negative and false positive rates are calculated with regard to a defect-free class only. By merging the two predicted classes of crater and dirt to form a single class of defects, we thus consider the binary case of defect versus defect-free.  The rationale behind doing so is that the main interest is in detecting defects rather than in distinguishing between crater (or dirt) and crater-free (or dirt-free) patches.

\section{Experimental results}\label{section:results}

\begin{figure}[H]
  \centering{}
  \includegraphics[width=0.7\textwidth]{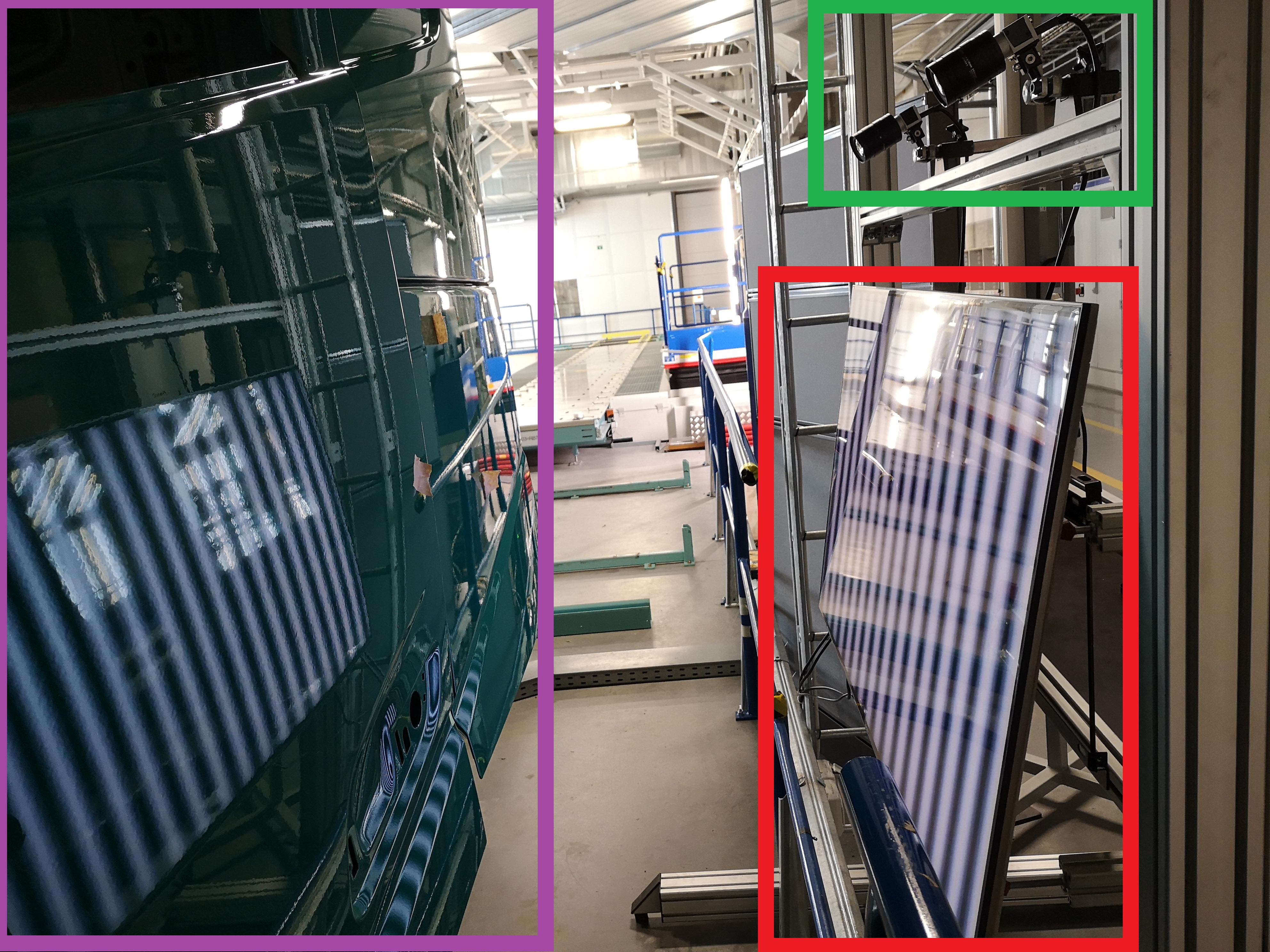}%{volvo-rig.jpg}
  
  \caption{\label{set-up screen camera}The setup of the test system.}
\end{figure}

\begin{figure}[H]
\centering
\begin{subfigure}{.35\textwidth}
  \centering
  \includegraphics[width=.7\linewidth]{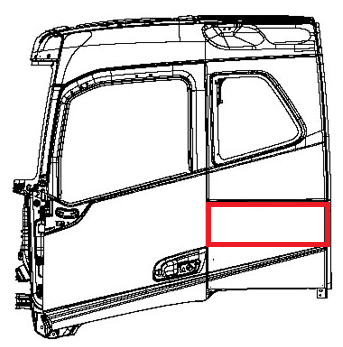}
  \caption{}
  \label{fig:FH-cab}
\end{subfigure}%
\begin{subfigure}{.35\textwidth}
  \centering
  \includegraphics[width=.7\linewidth]{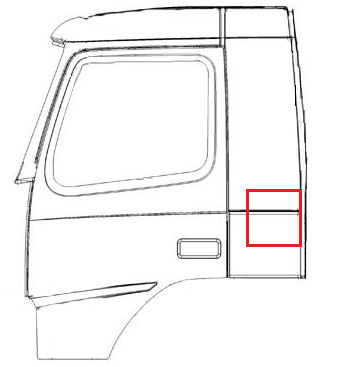}
  \caption{}
  \label{fig:FM-Long_Cab}
\end{subfigure}
\begin{subfigure}{.35\textwidth}
  \centering
  \includegraphics[width=.7\linewidth]{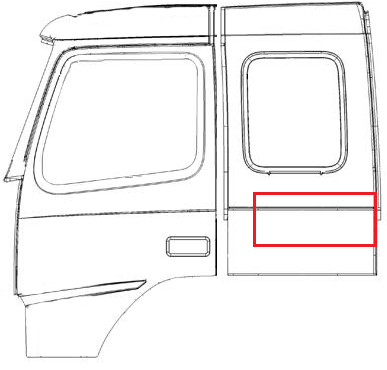}
  \caption{}
  \label{fig:FM-Small-Cab}
\end{subfigure}%
\caption{Sketches of the three inspected types of cabs. The target areas of the luggage lid surface are marked by red rectangles. (a) FH cab, (b) FM Small cab, and (c) FM Long cab.}
\label{fig:cabs}
\end{figure}

The proposed defect detection approach has been applied on images obtained from the pilot system installed at the paint shop of the Volvo GTO Cab plant located in Umeå, Sweden. Figure \ref{set-up screen camera} displays the setup of our pilot system. It consists of a 55 inch Sony screen (red rectangle in Figure \ref{set-up screen camera}) and two Fujinon HF16SA1 cameras (green rectangle). The inspected cabin is shown in a purple rectangle. The dimensions of the images taken from each of the two cameras are $2456 \times 2052$ with the sizes of around 5 megabits (Mb) each. The images from both cameras were combined to form one large image of about 9.6 Mb size and the dimensions of $4928 \times 2056$. The Sony screen has about 124cm of width, 72cm of height and 139cm of diagonal. The screen's resolution is $3840 \times 2160$ (4K display). The description of the geometrical measurements used in the setup of the machine vision system is skipped since it is of little importance here. 
%\todo[inline]{Describe, in caption, what the red, green and purple rectangles are displaying in figure \ref{set-up screen camera}./Blaise}
The surface of luggage lids of cab bodies was targeted by the considered pilot system. Three types of cabs in the production line were inspected, such as FH cabs, FM Small and FM Long cabs (see Figure \ref{fig:cabs}). The approximate sizes of the covered surfaces (red-marked rectangles in Figure \ref{fig:cabs}) were (a)
 $77\times 30$ cm$^2$ for the FH cabs, (b) $25\times 30$ cm$^2$ for the FM Small cabs, and (c) $62\times 30$ cm$^2$ for the FM Long cabs.

As a cab body goes through a complex multistage painting process, defects of various nature might appear on its surface. However, the most common defects are dirt and crater, thereby the focus of this work is specifically on these two types of defects. Dirt can be described as a small bump deposited in, on, or under the painted surface, whereas crater looks like a circular low spot or bowl-shaped cavity on the surface. 
It should be noted that it is rather difficult to distinguish dirt from crater, so human touch is often needed to identify those during the manual inspection. At the same time dirt defect is observed more frequently than crater on the cab bodies.

\subsection{Dataset}

The dataset, obtained by the deflectometry-based image acquisition approach described in Section \ref{section:deflectometry},  consists of patches of sixteen channels. The channels were formed by considering different combinations of the two parameters of the projected sinusoidal pattern, such as frequency parameter $f$ and phase parameter $\psi.$ They were specified as $f\in\{8, 16, 32, 64\}$ and $\psi\in \{0,\pi/2,\pi, 3\pi/2\}$.  This gives a possibility to consider sixteen subsets of patches, where each subset is composed of patches corresponding to a single channel (e.g., channel: $[f=8,\psi=0]$). In addition, for every channel patches of eight different sizes ($m\times m$) were extracted from the captured images, $m\in \{31,51,71,91,111,131,151,171\}$. 

In total, 18433 labeled patches of each size were collected, of those 4234 were labeled as dirt, 372 as craters, and 13827 as defect-free. The preliminary analysis has showed that the value of the phase parameter does not have significant effect on the classifier performance, so $\psi=\pi$ was chosen for further investigation. A trade-off between computational efficiency and classifier performance has led to the patch size $91\times 91.$ Therefore, we present the results on four subsets of patches that correspond to four channels that have the same value of phase, $\psi=\pi$, the same size of $91\times 91$, but four different values of frequency $f$.

\subsection{Binary classification problem}

In order to evaluate the performance of the proposed approach, we examine two types of the classification task: a binary task of classifying patches into defect/defect-free groups, and three-class task of classifying patches into crater/dirt/defect-free groups. In experiments, we focus on comparison of the proposed EDF features with several conventional feature descriptors, including histogram of oriented gradients (HOG) \citep{dalal05}, local binary pattern (LBP) \citep{ojala2002multiresolution}, 2d discrete wavelet transform (DWT) \citep{haar1910theorie,daubechies1992ten}, and features based on variability. Since the main advantage of the proposed features is in the usage of the straightforward probabilistic $k$-NN classifier, we employ this classifier when comparing features for both classification tasks, but in addition, apply support vector machines (SVM) for the binary class task. The choice of SVM was made on the grounds that during our preliminary study it showed better performance results among other examined classifiers including random forest and neural networks. All experiments were carried out in R 3.4.4 environment \citep{cranr}. The R library \verb+e1071+ \citep{e1071} was used to train a support vector machine, and \verb+knnx.dist()+ function of the R package \verb+FNN+ \citep{fnn} to calculate the Euclidean distances of one-nearest neighbors. 

The datasets were divided into training and validation sets, with 70\% of data used for training. Stratified sampling was applied to split the data. This assures that the amount of patches in each of the classes in the training/validation set is proportional to that of the total dataset. We employ  a five-fold cross validation technique to tune the SVM parameters. Furthermore, the stratified sampling was repeated five times for SVM, with the five-fold cross validation performed on every split, and ten times for probabilistic $k$-NN.      
Tables \ref{tab:svm-binary} and \ref{tab:knn-binary} show experimental results for the binary classification problem using SVM and probabilistic $k$-NN classifier, respectively. The performance of the proposed feature vector (denoted as `EDF' in the tables) was compared with the performances of the eight others: i) HOG, histogram of oriented gradients \citep{dalal05}; ii) LBP-HOG, HOG features fetched from measure for local image contrast \citep{ojala2002multiresolution}; iii) HOG:LBP-HOG, HOG and LBP-HOG features together; iv) HOG-Haar, HOG features derived from Haar wavelet's denoised patches \citep{haar1910theorie}; v) HOG-Daublets, Daublets wavelet's denoised patches based HOG features \citep{daubechies1992ten} vi) HOG-Symmlets, HOG features extracted from Symmlet wavelet's denoised patches \citep{daubechies1992ten}; vii) col.std, column standard deviations of the grey intensity pixel values; viii) col.std:HOG, column standard deviations and HOG feature descriptors jointly.
%\todo[inline,color=yellow]{Blaise, could you please give short descriptions for each fv you used with the corresponding references?} %\todo[inline]{Done it. /Blaise}

The classification results show that the proposed approach is very promising. In comparison with all considered feature descriptors, the proposed EDF based feature vector performs the best in terms of both conventional and probability based performance metrics. The value of the frequency parameter of the sinusoidal pattern has only marginal influence on the detection performance of the EDF. In contrast, the performance of the alternative features decreases for higher values of the frequency, which is even more profound when using the probabilistic $k$-NN (Table \ref{tab:knn-binary}). Furthermore, the proposed features showed equally good classification results in terms of false positive and false negative rates, whereas the competitive approaches appear less capable of detecting defect, erroneously yielding its absence. This indicates that the EDF features achieve not only higher classification performance but also can be viewed as an optimal approach to resolve the sensitivity-specificity trade-off task. Moreover, the introduced probability based evaluation metrics suggest that the approach is very certain in all its decisions made.

%\begin{table}[H]
%\small

\begin{sidewaystable}
\scriptsize
\caption{\label{tab:svm-binary} SVM results for binary classification problem for four channels. Performance results are means over five runs. The corresponding standard errors are given in brackets. The best rates are in bold.}
\centering
\begin{tabular}{|c||c|c|c|c|c|c|c|}
\hline
\multicolumn{8}{|c|}{{\color{orange}Channel: $[f=8,~\psi=\pi]$}} \\
\hline
Feature vector & Entropy & MER & FPR & FNR  & probMER & probFPR & probFNR  \\ 
\hline
EDF      & \tiny{${\bf 4.43\cdot 10^{-3}}(4.61\cdot 10^{-3})$} & \tiny{${\bf 7.95\cdot 10^{-4}}(3.28\cdot 10^{-4})$} & \tiny{${\bf 6.76\cdot 10^{-4}}(3.15\cdot 10^{-4})$} & 
\tiny{${\bf 1.14\cdot 10^{-3}}(1.48\cdot 10^{-3})$}
   & \tiny{${\bf 1.64\cdot 10^{-3}}(2.29\cdot 10^{-4})$}  & \tiny{${\bf 1.29\cdot 10^{-3}}(2.61\cdot 10^{-4})$}  &  \tiny{${\bf 2.68\cdot 10^{-3}}(1.11\cdot 10^{-3})$} \\
\hline
HOG & \tiny{$0.165(3.7\cdot 10^{-3})$}  & \tiny{$0.05(2.9\cdot 10^{-3})$}  & \tiny{$0.01(2.6\cdot 10^{-3})$}  &\tiny{$0.17(9\cdot 10^{-3})$}  &\tiny{$0.087(3\cdot 10^{-3})$}  & \tiny{$0.059(2.4\cdot 10^{-3})$}  & \tiny{$0.17(9.7\cdot 10^{-3})$}   \\
\hline
LBP-HOG  &\tiny{$0.52(2.3\cdot 10^{-3})$}  & \tiny{$0.24(2.2\cdot 10^{-3})$}  & \tiny{$0.02(7.7\cdot 10^{-3})$} & \tiny{$0.92(0.03)$}  & \tiny{$0.35(2\cdot 10^{-3})$} & \tiny{$0.23(3.4\cdot 10^{-3})$}  & \tiny{$0.70(0.015)$}   \\
\hline
HOG : LBP-HOG & \tiny{$0.17(4.7\cdot 10^{-3})$}  & \tiny{$0.05(3.3\cdot 10^{-3})$}  & \tiny{$8 \cdot 10^{-3} (2 \cdot 10^{-3})$} & \tiny{$0.17(0.01)$}   & \tiny{$0.09(2.8 \cdot 10^{-3})$}  & \tiny{$0.06(1.9\cdot 10^{-3})$} & \tiny{$0.18(9\cdot 10^{-3})$}   \\
\hline
HOG-Haar  &\tiny{$0.15(4\cdot 10^{-3})$}  & \tiny{$0.046(3\cdot 10^{-3})$}  &\tiny{$0.01(1.2\cdot 10^{-3})$}  &\tiny{$0.15(0.01)$}  &\tiny{$0.076(2.1\cdot 10^{-3})$}  &\tiny{$0.05(2.2\cdot 10^{-3})$}  & \tiny{$0.15(1\cdot 10^{-3})$}  \\
\hline
HOG-Daublets  & \tiny{$0.13(3.2\cdot 10^{-3})$} & \tiny{$0.041(3.6\cdot 10^{-3})$}  & \tiny{$9.4\cdot 10^{-3}(1.9\cdot 10^{-3})$}  & \tiny{$0.14(0.01)$}  & \tiny{$0.072(2.6\cdot 10^{-3})$} & \tiny{$0.048(1.6\cdot 10^{-3})$}  & \tiny{$0.14(9.6\cdot 10^{-3})$}   \\
\hline
HOG-Symmlets  & \tiny{$0.14(5.5\cdot 10^{-3})$} & \tiny{$0.043(2.9\cdot 10^{-3})$}  & \tiny{$0.01(2.6\cdot 10^{-3})$}  & \tiny{$0.14(9.7\cdot 10^{-3})$}  & \tiny{$0.07(2.2\cdot 10^{-3})$}  & \tiny{$0.048(2.8\cdot 10^{-3})$}  & \tiny{$0.14(0.01)$}  \\
\hline
col.std & \tiny{$0.254(3.06\cdot 10^{-3})$} & \tiny{$0.085(3.76\cdot 10^{-3})$} & \tiny{$0.023(2.08\cdot 10^{-3})$} & \tiny{$0.267(9.50\cdot 10^{-3})$} & \tiny{$0.141(2.56\cdot 10^{-3})$} & \tiny{$0.095(1.94\cdot 10^{-3})$} & \tiny{$0.278(7.49\cdot 10^{-3})$}  \\
\hline
%P-splines : col.std & \tiny{$0.21(3.71\cdot 10^{-3})$} & \tiny{$0.073(3.49\cdot 10^{-3})$} & \tiny{$0.022(2.01\cdot 10^{-3})$} & \tiny{$0.226(0.013)$} & \tiny{$0.117(1.59\cdot 10^{-3})$} & \tiny{$0.077(2.49\cdot 10^{-3})$} & \tiny{$0.234(9.20\cdot 10^{-3})$}  \\
%\hline
col.std : HOG & \tiny{$0.184(1.46\cdot 10^{-3})$} & \tiny{$0.058(2.74\cdot 10^{-3})$} & \tiny{$0.015(2.04\cdot 10^{-3})$} & \tiny{$0.184(8.90\cdot 10^{-3})$} & \tiny{$0.099(2.36\cdot 10^{-3})$} &\tiny{$0.065(1.94\cdot 10^{-3})$}  & \tiny{$0.198(8.95\cdot 10^{-3})$}  \\
\hline
\multicolumn{8}{|c|}{{\color{orange}Channel: $[f=16,~\psi=\pi]$}} \\
\hline
Feature vector & Entropy & MER & FPR & FNR  & probMER & probFPR & probFNR  \\
\hline
   EDF      & \tiny{${\bf 0.010}(1.11\cdot 10^{-3})$} & \tiny{${\bf 2.28\cdot 10^{-3}}(6.46\cdot 10^{-4})$} & \tiny{${\bf 1.64\cdot 10^{-3}}(3.97\cdot 10^{-4})$} & \tiny{${\bf 4.15\cdot 10^{-3}}(1.85\cdot 10^{-3})$} & \tiny{${\bf 4.31\cdot 10^{-3}}(4.80\cdot 10^{-4})$} & \tiny{${{\bf 3.29\cdot 10^{-3}}(4.42\cdot 10^{-4})}$} & 
  \tiny{${\bf 7.33\cdot 10^{-3}}(1.55\cdot 10^{-3})$} \\
\hline
HOG & \tiny{$0.16(3\cdot 10^{-3})$}   & \tiny{$0.05(2.6\cdot 10^{-3})$}  & \tiny{$0.01(1.9\cdot 10^{-3})$}    & \tiny{$0.16(0.01)$} & \tiny{$0.08(3\cdot 10^{-3})$}  & \tiny{$0.06(2.4\cdot 10^{-3})$}  & \tiny{$0.17(0.01)$}   \\
\hline
LBP-HOG  & \tiny{$0.52(0.01)$}  & \tiny{$0.24(4.5\cdot 10^{-3})$}  & \tiny{$0.023(6.7\cdot 10^{-3})$}  & \tiny{$0.90(0.034)$}  & \tiny{$0.35(7.7\cdot 10^{-3})$}  & \tiny{$0.23(7\cdot 10^{-3})$}  & \tiny{$0.70(0.01)$}   \\
\hline
HOG : LBP-HOG & \tiny{$0.16(3\cdot 10^{-3})$}  & \tiny{$0.049(2.6\cdot 10^{-3})$}  & \tiny{$0.01(1.2\cdot 10^{-3})$}  & \tiny{$0.16(8.4\cdot 10^{-3})$}  & \tiny{$0.087(3\cdot 10^{-3})$}  & \tiny{$0.058(2.3\cdot 10^{-3})$}  & \tiny{$0.17(8.7\cdot 10^{-3})$}  \\
\hline
HOG-Haar  & \tiny{$0.15(3\cdot 10^{-3})$}  & \tiny{$0.05(3.4\cdot 10^{-3})$}  & \tiny{$0.016(3.8\cdot 10^{-3})$}  & \tiny{$0.15(0.01)$}  & \tiny{$0.08(3.2\cdot 10^{-3})$}  & \tiny{$0.06(2.9\cdot 10^{-3})$}  & \tiny{$0.16(0.01)$}   \\
\hline
HOG-Daublets  & \tiny{$0.14(3.4\cdot 10^{-3})$}  & \tiny{$0.048(2.5\cdot 10^{-3})$}  & \tiny{$0.014(2.3\cdot 10^{-3})$}  & \tiny{$0.15(0.01)$}  & \tiny{$0.08(1.6\cdot 10^{-3})$}  & \tiny{$0.05(2.3\cdot 10^{-3})$}  & \tiny{$0.16(0.01)$}   \\
\hline
HOG-Symmlets  & \tiny{$0.14(3.5\cdot 10^{-3})$}  & \tiny{$0.047(3.9\cdot 10^{-3})$}  & \tiny{$0.01(2\cdot 10^{-3})$}  & \tiny{$0.15(0.01)$}  & \tiny{$0.08(2\cdot 10^{-3})$}  & \tiny{$0.05(2.3\cdot 10^{-3})$}  & \tiny{$0.15(8.9\cdot 10^{-3})$}   \\
\hline
col.std & \tiny{$0.331(3.79\cdot 10^{-3})$} & \tiny{$0.126(3.34\cdot 10^{-3})$} & \tiny{$0.037(5.13\cdot 10^{-3})$} & \tiny{$0.390(3.83\cdot 10^{-3})$}  & \tiny{$0.197(1.22\cdot 10^{-3})$} & \tiny{$0.131(2.24\cdot 10^{-3})$} & \tiny{$0.391(2.96\cdot 10^{-3})$}  \\
\hline
%P-splines : col.std & \tiny{$0.278(2.43\cdot 10^{-3})$} &  \tiny{$0.107(2.89\cdot 10^{-3})$} &  \tiny{$0.034(4.09\cdot 10^{-3})$} &  \tiny{$0.321(5.70\cdot 10^{-3})$}  &  \tiny{$0.162(7.05\cdot 10^{-4})$} &  \tiny{$0.109(1.34\cdot 10^{-3})$}  &   \tiny{$0.322(4.12\cdot 10^{-3})$} \\
%\hline
col.std : HOG & \tiny{$0.181(2.64\cdot 10^{-3})$}  & \tiny{$0.061(1.88\cdot 10^{-3})$} & \tiny{$0.019(2.65\cdot 10^{-3})$} & \tiny{$0.187(9.44\cdot 10^{-3})$} & \tiny{$0.101(1.22\cdot 10^{-3})$} & \tiny{$0.066(2.22\cdot 10^{-3})$} & \tiny{$0.203(5.95\cdot 10^{-3})$}  \\
\hline
\multicolumn{8}{|c|}{{\color{orange}Channel: $[f=32,~\psi=\pi]$}} \\
\hline
Feature vector & Entropy & MER & FPR & FNR  & probMER & probFPR & probFNR  \\
\hline
  EDF      & \tiny{${\bf 7.32\cdot 10^{-3}}(8.23\cdot 10^{-4})$}  & \tiny{${\bf 1.45 \cdot 10^{-3}}(6.88\cdot 10^{-4})$} & \tiny{${\bf 6.77\cdot 10^{-4}}(3.97\cdot 10^{-4})$} & \tiny{${\bf 3.72\cdot 10^{-3}}(2.05\cdot 10^{-3})$} & \tiny{${\bf 3.02\cdot 10^{-3}}(7.30\cdot 10^{-4})$} & \tiny{${\bf 2.02\cdot 10^{-3}}(4.04\cdot 10^{-4})$} &  \tiny{${\bf 5.98\cdot 10^{-3}}(2.43\cdot 10^{-3})$} \\
\hline
HOG & \tiny{$0.19(4\cdot 10^{-3})$}  & \tiny{$0.06(4\cdot 10^{-3})$}   & \tiny{$0.02(3.9\cdot 10^{-3})$}   & \tiny{$0.20(0.02)$}   & \tiny{$0.11(3.4\cdot 10^{-3})$}   & \tiny{$0.07(2\cdot 10^{-3})$}  & \tiny{$0.21(0.01)$}  \\
\hline
LBP-HOG  & \tiny{$0.49(0.04)$}  & \tiny{$0.25(2.2\cdot 10^{-3})$}  & \tiny{$0.014(8.6\cdot 10^{-3})$}  & \tiny{$0.95(0.03)$}  & \tiny{$0.34(0.016)$}  & \tiny{$0.21(0.036)$}  & \tiny{$0.74(0.046)$}  \\
\hline
HOG : LBP-HOG & \tiny{$0.20(6.4\cdot 10^{-3})$}  & \tiny{$0.065(4\cdot 10^{-3})$}  & \tiny{$0.02(2.3\cdot 10^{-3})$}  & \tiny{$0.20(0.015)$}  & \tiny{$0.11(3.3\cdot 10^{-3})$}  & \tiny{$0.073(3.2\cdot 10^{-3})$}  & \tiny{$0.22(0.012)$}   \\
\hline
HOG-Haar  & \tiny{$0.21(3\cdot 10^{-3})$}  & \tiny{$0.07(3.6\cdot 10^{-3})$}  & \tiny{$0.03(4\cdot 10^{-3})$}  & \tiny{$0.21(0.014)$}  & \tiny{$0.12(3.2\cdot 10^{-3})$}  & \tiny{$0.08(3.7\cdot 10^{-3})$}  & \tiny{$0.23(0.013)$}   \\
\hline
HOG-Daublets  & \tiny{$0.19(3.8\cdot 10^{-3})$}  & \tiny{$0.07(4\cdot 10^{-3})$}  & \tiny{$0.026(5\cdot 10^{-3})$}  & \tiny{$0.20(0.013)$}  & \tiny{$0.11(3.1\cdot 10^{-3})$}  & \tiny{$0.07(4\cdot 10^{-3})$}  & \tiny{$0.22(0.013)$}   \\
\hline
HOG-Symmlets  & \tiny{$0.19(1.4\cdot 10^{-3})$}  & \tiny{$0.069(3.9\cdot 10^{-3})$}  & \tiny{$0.03(4.4\cdot 10^{-3})$}  & \tiny{$0.20(0.017)$}  & \tiny{$0.11(3.2\cdot 10^{-3})$}  & \tiny{$0.07(3.8\cdot 10^{-3})$}  & \tiny{$0.21(0.016)$}   \\
\hline
col.std & \tiny{$0.373(4.65\cdot 10^{-3})$} & \tiny{$0.149(2.71\cdot 10^{-3})$} & \tiny{$0.037(3.85\cdot 10^{-3})$} & \tiny{$0.481(0.015)$} & \tiny{$0.227(2.05\cdot 10^{-3})$} & \tiny{$0.151(3.37\cdot 10^{-3})$} & \tiny{$0.451(9.77\cdot 10^{-3})$}  \\
\hline
%P-splines : col.std & \tiny{$0.333(3.51\cdot 10^{-3})$} & \tiny{$0.129(4.05\cdot 10^{-3})$} & \tiny{$0.035(2.54\cdot 10^{-3})$} & \tiny{$0.410(0.014)$} & \tiny{$0.200(2.24\cdot 10^{-3})$} & \tiny{$0.133(2.67\cdot 10^{-3})$} & \tiny{$0.400(9.89\cdot 10^{-3})$}  \\
%\hline
col.std : HOG & \tiny{$0.221(2.21\cdot 10^{-3})$} & \tiny{$0.080(1.66\cdot 10^{-3})$} & \tiny{$0.027(2.59\cdot 10^{-3})$} & \tiny{$0.236(4.51\cdot 10^{-3})$} &\tiny{$0.127(1.68\cdot 10^{-3})$}  & \tiny{$0.084(2.12\cdot 10^{-3})$} & \tiny{$0.254(3.71\cdot 10^{-3})$}  \\
\hline
\multicolumn{8}{|c|}{{\color{orange}Channel: $[f=64,~\psi=\pi]$}} \\
\hline
Feature vector & Entropy & MER & FPR & FNR  & probMER & probFPR & probFNR  \\ 
\hline
EDF      & \tiny{${\bf 6.69\cdot 10^{-3}}(1.39\cdot 10^{-3})$} & \tiny{${\bf 7.23\cdot 10^{-5}}(9.89\cdot 10^{-5})$} & \tiny{${\bf 4.83\cdot 10^{-5}}(1.08\cdot 10^{-4})$} & \tiny{${\bf 1.43\cdot 10^{-4}}(3.20\cdot 10^{-4})$} & \tiny{${\bf 1.96\cdot 10^{-3}}(4.51\cdot 10^{-4})$} & \tiny{${\bf 2.28\cdot 10^{-3}}(6.11\cdot 10^{-4})$} & \tiny{${\bf 1.02\cdot 10^{-3}}(5.83\cdot 10^{-4})$}  \\
\hline
HOG & \tiny{$0.29(6\cdot 10^{-3})$}  & \tiny{$0.12(4\cdot 10^{-3})$}  & \tiny{$0.04(3.6\cdot 10^{-3})$}   & \tiny{$0.33(0.014)$}  & \tiny{$0.17(1.8\cdot 10^{-3})$}  & \tiny{$0.12(1.3\cdot 10^{-3})$}  & \tiny{$0.35(0.01)$}   \\
\hline
LBP-HOG  & \tiny{$0.53(0.01)$}  & \tiny{$0.25(1.3\cdot 10^{-3})$}  & \tiny{$6 \cdot 10^{-3}(2.6\cdot 10^{-3})$}  & \tiny{$0.96(0.012)$}  & \tiny{$0.35(4.7\cdot 10^{-3})$}  & \tiny{$0.23(9.4\cdot 10^{-3})$}  & \tiny{$0.73(0.01)$}   \\
\hline
HOG : LBP-HOG & \tiny{$0.3(5\cdot 10^{-3})$}  & \tiny{$0.12(5.4\cdot 10^{-3})$}  &\tiny{$0.04(3.3\cdot 10^{-3})$}  & \tiny{$0.34(0.021)$}  & \tiny{$0.18(3.9\cdot 10^{-3})$}  & \tiny{$0.12(1.8\cdot 10^{-3})$}  & \tiny{$0.36(0.014)$}   \\
\hline
HOG-Haar  & \tiny{$0.32(6.2\cdot 10^{-3})$}  & \tiny{$0.13(5.2\cdot 10^{-3})$}  & \tiny{$0.05(3.9\cdot 10^{-3})$}  & \tiny{$0.37(0.016)$}  & \tiny{$0.19(2.5\cdot 10^{-3})$}  & \tiny{$0.13(1.4\cdot 10^{-3})$}  & \tiny{$0.39(8.7\cdot 10^{-3})$}  \\
\hline
HOG-Daublets  & \tiny{$0.28(0.023)$}  & \tiny{$0.12(4.9\cdot 10^{-3})$}  & \tiny{$0.04(8\cdot 10^{-3})$}  & \tiny{$0.35(0.037)$}  &\tiny{$0.17(5.9\cdot 10^{-3})$}  & \tiny{$0.11(0.017)$}  & \tiny{$0.37(0.03)$}   \\
\hline
HOG-Symmlets  & \tiny{$0.30(5\cdot 10^{-3})$}  & \tiny{$0.12(4.9\cdot 10^{-3})$}  & \tiny{$0.04(1\cdot 10^{-3})$}  & \tiny{$0.33(0.019)$}  & \tiny{$0.18(2\cdot 10^{-3})$}  & \tiny{$0.12(1.8\cdot 10^{-3})$}  & \tiny{$0.35(0.012)$}   \\
\hline
col.std & \tiny{$0.441(2.77\cdot 10^{-3})$} & \tiny{$0.194(4.04\cdot 10^{-3})$} & \tiny{$0.050(3.69\cdot 10^{-3})$} & \tiny{$0.619(6.87\cdot 10^{-3})$} & \tiny{$0.280(1.05\cdot 10^{-3})$} & \tiny{$0.187(2.23\cdot 10^{-3})$} & \tiny{$0.555(3.27\cdot 10^{-3})$}  \\
\hline
%P-splines : col.std & \tiny{$0.420(5.42\cdot 10^{-3})$} & \tiny{$0.187(1.10\cdot 10^{-3})$} & \tiny{$0.059(2.79\cdot 10^{-3})$} & \tiny{$0.565(5.60\cdot 10^{-3})$} & \tiny{$0.266(1.50\cdot 10^{-3})$} & \tiny{$0.178(2.07\cdot 10^{-3})$} & \tiny{$0.527(2.47\cdot 10^{-3})$}  \\
%\hline
col.std : HOG & \tiny{$0.305(3.36\cdot 10^{-3})$} & \tiny{$0.122(1.68\cdot 10^{-3})$} & \tiny{$0.047(2.34\cdot 10^{-3})$} & \tiny{$0.346(5.80\cdot 10^{-3})$} &\tiny{$0.185(1.27\cdot 10^{-3})$}  & \tiny{$0.125(2.35\cdot 10^{-3})$} & \tiny{$0.361(6.14\cdot 10^{-3})$}  \\
\hline
\end{tabular}
%\end{table}
\end{sidewaystable}

%\begin{table}[H]

%\scriptsize
\begin{sidewaystable}
\scriptsize
\caption{\label{tab:knn-binary} Probabilistic $k$-NN results for binary classification problem for four channels. Performance results are means over ten runs. The corresponding standard errors are given in brackets.}
\centering
\begin{tabular}{|c||c|c|c|c|c|c|c|}
\hline
\multicolumn{8}{|c|}{{\color{orange}Channel: $[f=8,~\psi=\pi]$}} \\
\hline
Feature vector & Entropy & MER & FPR & FNR  & probMER & probFPR & probFNR  \\ 
\hline
EDF      & \tiny{${\bf 1.42\cdot 10^{-6}}(1.73\cdot 10^{-6})$} & \tiny{${\bf\approx 0}(0)$} &  \tiny{${\bf\approx 0}(0)$} & \tiny{${\bf\approx 0}(0)$} & \tiny{${\bf 1.74\cdot 10^{-7}}(2.33\cdot 10^{-7})$}  & 
\tiny{${\bf 2.21\cdot 10^{-7}}(3.17\cdot 10^{-7})$} & \tiny{${\bf 3.42\cdot 10^{-8}}(8.17\cdot 10^{-8})$}  \\
\hline
%edf(unscaled)      & \tiny{$2.10\cdot 10^{-7}(3.07\cdot 10^{-7})$} & \tiny{$0(0)$} &  \tiny{$0(0)$} & \tiny{$0(0)$} & \tiny{$2.15\cdot 10^{-8}(3.50\cdot 10^{-8})$}  & \tiny{$1.08\cdot 10^{-8}(1.72\cdot 10^{-8})$} & \tiny{$5.32\cdot 10^{-8}(1.31\cdot 10^{-7})$}  \\
%\hline
HOG & \tiny{$0.074(2.9\cdot 10^{-3})$} &\tiny{$0.22(3.7\cdot 10^{-3})$}  & \tiny{$0.11(3.9\cdot 10^{-3})$}  & \tiny{$0.55(0.011)$}  & \tiny{$0.22(3.5\cdot 10^{-3})$} &\tiny{$0.12(3.9\cdot 10^{-3})$}  & \tiny{$0.55(8.9\cdot 10^{-3})$}  \\
\hline
LBP-HOG  & \tiny{$0.27(3.3\cdot 10^{-3})$} & \tiny{$0.35(5.6\cdot 10^{-3})$}  & \tiny{$0.23(4.7\cdot 10^{-3})$}  & \tiny{$0.73(0.013)$} & \tiny{$0.37(3.9\cdot 10^{-3})$}  &\tiny{$0.26(3.2\cdot 10^{-3})$}  & \tiny{$0.70(9.6\cdot 10^{-3})$}   \\
\hline
HOG:LBP-HOG & \tiny{$0.15(3.3\cdot 10^{-3})$}  & \tiny{$0.35(5.5\cdot 10^{-3})$}  & \tiny{$0.23(4.7\cdot 10^{-3})$}  & \tiny{$0.73(0.013)$}  & \tiny{$0.36(4.6\cdot 10^{-3})$}  & \tiny{$0.24(3.5\cdot 10^{-3})$}  & \tiny{$0.72(0.01)$}  \\
\hline
HOG-Haar  & \tiny{$0.06(2\cdot 10^{-3})$}  & \tiny{$0.21(3.6\cdot 10^{-3})$}  & \tiny{$0.10(4.3\cdot 10^{-3})$}  & \tiny{$0.52(9.2\cdot 10^{-3})$}  & \tiny{$0.21(3.7\cdot 10^{-3})$}  & \tiny{$0.10(4.2\cdot 10^{-3})$}  & \tiny{$0.52(8.6\cdot 10^{-3})$}  \\
\hline
HOG-Daublets  & \tiny{$0.07(2\cdot 10^{-3})$}   & \tiny{$0.22(3\cdot 10^{-3})$}   & \tiny{$0.11(4.9\cdot 10^{-3})$}   & \tiny{$0.54(0.01)$}   & \tiny{$0.22(2.7\cdot 10^{-3})$}   & \tiny{$0.11(4\cdot 10^{-3})$}   & \tiny{$0.54(0.01)$}   \\
\hline
HOG-Symmlets  & \tiny{$0.07(1.6\cdot 10^{-3})$}    & \tiny{$0.22(3.8\cdot 10^{-3})$}   & \tiny{$0.11(6\cdot 10^{-3})$}    & \tiny{$0.53(0.013)$}    & \tiny{$0.22(3.8\cdot 10^{-3})$}    & \tiny{$0.11(5\cdot 10^{-3})$}    & \tiny{$0.53(0.01)$}     \\
\hline
col.std & \tiny{$0.081(1.48\cdot 10^{-3})$} & \tiny{$0.136(1.92\cdot 10^{-3})$} & \tiny{$0.105(3.42\cdot 10^{-3})$} & \tiny{$0.226(7.69\cdot 10^{-3})$}  & \tiny{$0.143(2.28\cdot 10^{-3})$}  & \tiny{$0.116(3.74\cdot 10^{-3})$}  & \tiny{$0.225(6.51\cdot 10^{-3})$}     \\
\hline
%P-splines & \tiny{$0.228(4.56\cdot 10^{-3})$} & \tiny{$0.262(4.30\cdot 10^{-3})$}  &  \tiny{$0.130(4.17\cdot 10^{-3})$} &    \tiny{$0.652(0.0136)$} & \tiny{$0.279(3.84\cdot 10^{-3})$} & \tiny{$0.159(4.40\cdot 10^{-3})$} & \tiny{$0.633(0.010)$}  \\
%\hline
%P-splines:col.std & \tiny{$0.035(1.35\cdot 10^{-3})$} & \tiny{$0.143(1.99\cdot 10^{-3})$} & \tiny{$0.053(2.57\cdot 10^{-3})$} & \tiny{$0.407(8.84\cdot 10^{-2})$} & \tiny{$0.144(1.86\cdot 10^{-3})$} & \tiny{$0.055(2.39\cdot 10^{-3})$} & \tiny{$0.406(8.09\cdot 10^{-3})$}  \\
%\hline
col.std:HOG & \tiny{$0.041(1.79\cdot 10^{-3})$} & \tiny{$0.182( 0.012)$} & \tiny{$0.155(0.016)$}  & \tiny{$0.261(8.57\cdot 10^{-3})$} & \tiny{$0.183(0.012)$} & \tiny{$0.157(0.015)$}  &  \tiny{$0.261(7.38\cdot 10^{-3})$} \\
\hline
\multicolumn{8}{|c|}{{\color{orange}Channel: $[f=16,~\psi=\pi]$}} \\
\hline
Feature vector & Entropy & MER & FPR & FNR  & probMER & probFPR & probFNR  \\
\hline
 EDF     & \tiny{${\bf 1.93\cdot 10^{-5}}(3.35\cdot 10^{-5})$}  & \tiny{${\bf \approx 0}(0)$} &  \tiny{${\bf\approx 0}(0)$} & \tiny{${\bf\approx 0}(0)$}  & 
          \tiny{${\bf 6.41\cdot 10^{-6}}(1.57\cdot 10^{-5})$} & \tiny{${\bf 7.06\cdot 10^{-6}}(2.12\cdot 10^{-5})$} &  
           \tiny{${\bf 4.48\cdot 10^{-6}}(7.92\cdot 10^{-6})$} \\
\hline
%edf(unscaled)      & \tiny{$3.39\cdot 10^{-6}(3.21\cdot 10^{-6})$} & \tiny{$0(0)$} &  \tiny{$0(0)$} & \tiny{$0(0)$} & \tiny{$4.41\cdot 10^{-7}(4.74\cdot 10^{-7})$}  & \tiny{$1.78\cdot 10^{-7}(3.70\cdot 10^{-7})$} & \tiny{$1.22\cdot 10^{-6}(1.87\cdot 10^{-6})$}  \\ \hline
HOG &\tiny{$0.085(1.4\cdot 10^{-3})$}  & \tiny{$0.23(5.6\cdot 10^{-3})$} &\tiny{$0.12(6.3\cdot 10^{-3})$}  & \tiny{$0.57(0.014)$}  &\tiny{$0.24(5.1\cdot 10^{-3})$}  & \tiny{$0.13(5.5\cdot 10^{-3})$} &\tiny{$0.57(0.014)$}   \\
\hline
LBP-HOG  & \tiny{$0.26(2.6\cdot 10^{-3})$}   & \tiny{$0.34(3\cdot 10^{-3})$}  & \tiny{$0.21(4.1\cdot 10^{-3})$}  & \tiny{$0.71(7.6\cdot 10^{-3})$}  & \tiny{$0.35(2.4\cdot 10^{-3})$}  & \tiny{$0.24(2.9\cdot 10^{-3})$}  & \tiny{$0.69(7\cdot 10^{-3})$}   \\
\hline
HOG:LBP-HOG & \tiny{$0.14(2.6\cdot 10^{-3})$}   & \tiny{$0.34(3\cdot 10^{-3})$}   & \tiny{$0.21(4.2\cdot 10^{-3})$}   & \tiny{$0.72(7.6\cdot 10^{-3})$}   & \tiny{$0.34(2.8\cdot 10^{-3})$}   & \tiny{$0.22(3.3\cdot 10^{-3})$}  & \tiny{$0.71(7.7\cdot 10^{-3})$}    \\
\hline
HOG-Haar & \tiny{$0.08(2.1\cdot 10^{-3})$}  & \tiny{$0.23(4.3\cdot 10^{-3})$}  & \tiny{$0.11(4.6\cdot 10^{-3})$}  & \tiny{$0.56(6.8\cdot 10^{-3})$}  & \tiny{$0.23(4\cdot 10^{-3})$}  & \tiny{$0.12(4.3\cdot 10^{-3})$}  & \tiny{$0.56(7.8\cdot 10^{-3})$}   \\
\hline
HOG-Daublets  & \tiny{$0.09(3.6\cdot 10^{-3})$}  & \tiny{$0.24(4.2\cdot 10^{-3})$}  & \tiny{$0.13(4.4\cdot 10^{-3})$}  & \tiny{$0.60(7.3\cdot 10^{-3})$} & \tiny{$0.25(4.6\cdot 10^{-3})$}  & \tiny{$0.13(4.4\cdot 10^{-3})$}  & \tiny{$0.60(7.1\cdot 10^{-3})$}   \\
\hline
HOG-Symmlets  & \tiny{$0.09(3.1\cdot 10^{-3})$} & \tiny{$0.24(2.3\cdot 10^{-3})$}  & \tiny{$0.12(3\cdot 10^{-3})$}  & \tiny{$0.58(0.013)$}  & \tiny{$0.24(2.4\cdot 10^{-3})$} & \tiny{$0.13(2.9\cdot 10^{-3})$}  & \tiny{$0.58(0.01)$}   \\
\hline
col.std & \tiny{$0.065(1.05\cdot 10^{-3})$} & \tiny{$0.197(1.76\cdot 10^{-3})$} & \tiny{$0.126(3.86\cdot 10^{-3})$}  & \tiny{$0.408(0.013)$} & \tiny{$0.198(2.53\cdot 10^{-3})$} & \tiny{$0.128(3.28\cdot 10^{-3})$} & \tiny{$0.406(0.012)$}  \\
\hline
%P-splines & \tiny{$0.183(3.48\cdot 10^{-3})$} & \tiny{$0.291(4.08\cdot 10^{-3})$} & \tiny{$0.151(6.14\cdot 10^{-3})$} &  \tiny{$0.705(9.61\cdot 10^{-3})$} & \tiny{$0.300(2.99\cdot 10^{-3})$} & \tiny{$0.169(5.31\cdot 10^{-3})$} &  \tiny{$0.689(8.51\cdot 10^{-3})$} \\ \hline
%P-splines:col.std & \tiny{$0.041(1.51\cdot 10^{-3})$} & \tiny{$0.192(5.68\cdot 10^{-3})$} & \tiny{$0.081(5.09\cdot 10^{-3})$} & \tiny{$0.521(0.014)$} & \tiny{$0.194(5.23\cdot 10^{-3})$} & \tiny{$0.083(4.87\cdot 10^{-3})$} & \tiny{$0.521(0.013)$}  \\\hline
col.std:HOG & \tiny{$0.037(1.8\cdot 10^{-3})$} & \tiny{$0.189(4.71\cdot 10^{-3})$} & \tiny{$0.100(4.28\cdot 10^{-3})$} & \tiny{$ 0.451(0.011)$} & \tiny{$0.190(5.02\cdot 10^{-3})$} & \tiny{$0.101(4.32\cdot 10^{-3})$} & \tiny{$0.452(0.011)$}  \\
\hline
\multicolumn{8}{|c|}{{\color{orange}Channel: $[f=32,~\psi=\pi]$}} \\
\hline
Feature vector & Entropy & MER & FPR & FNR  & probMER & probFPR & probFNR  \\
\hline
  EDF      & \tiny{${\bf 9.32\cdot 10^{-6}}(1.43\cdot 10^{-5})$} &     \tiny{${\bf\approx 0}(0)$} &  \tiny{${\bf\approx 0}(0)$} & \tiny{${\bf\approx 0}(0)$}  &
          \tiny{${\bf 2.06\cdot 10^{-6}}(4.04\cdot 10^{-6})$} & \tiny{${\bf 2.27\cdot 10^{-6}}(5.52\cdot 10^{-6})$} & 
           \tiny{${\bf 1.45\cdot 10^{-6}}(2.36\cdot 10^{-6})$} \\
\hline
%edf(unscaled)      & \tiny{$3.59 \cdot 10^{-6}(8.27\cdot 10^{-6})$} & \tiny{$0(0)$} &  \tiny{$0(0)$} & \tiny{$0(0)$} & \tiny{$7.06\cdot 10^{-7}(1.85\cdot 10^{-6})$}  & \tiny{$8.20\cdot 10^{-7}(2.46\cdot 10^{-6})$} & \tiny{$3.71\cdot 10^{-7}(4.55\cdot 10^{-7})$}  \\ \hline
HOG &\tiny{$0.12(2.1\cdot 10^{-3})$}  & \tiny{$0.27(3.7\cdot 10^{-3})$}  & \tiny{$0.15(5.3\cdot 10^{-3})$}  & \tiny{$0.66(1.3\cdot 10^{-3})$}  &\tiny{$0.28(3.2\cdot 10^{-3})$}  & \tiny{$0.15(4.3\cdot 10^{-3})$}  & \tiny{$0.65(0.012)$}   \\
\hline
LBP-HOG  & \tiny{$0.27(1.9\cdot 10^{-3})$}  & \tiny{$0.35(3.4\cdot 10^{-3})$}  & \tiny{$0.22(4\cdot 10^{-3})$}  & \tiny{$0.75(8.3\cdot 10^{-3})$}  & \tiny{$0.37(3\cdot 10^{-3})$}  & \tiny{$0.25(3.6\cdot 10^{-3})$}  & \tiny{$0.72(8\cdot 10^{-3})$}  \\
\hline
HOG:LBP-HOG & \tiny{$0.14(1.7\cdot 10^{-3})$} & \tiny{$0.35(3.4\cdot 10^{-3})$}   & \tiny{$0.22(4\cdot 10^{-3})$}   & \tiny{$0.75(8.3\cdot 10^{-3})$}   & \tiny{$0.36(3.3\cdot 10^{-3})$}   & \tiny{$0.23(4.2\cdot 10^{-3})$}   & \tiny{$0.74(8.8\cdot 10^{-3})$}    \\
\hline
HOG-Haar  & \tiny{$0.11(1.8\cdot 10^{-3})$} & \tiny{$0.26(2.9\cdot 10^{-3})$}  & \tiny{$0.14(1.9\cdot 10^{-3})$} & \tiny{$0.62(9.6\cdot 10^{-3})$}  & \tiny{$0.26(3.4\cdot 10^{-3})$}  & \tiny{$0.15(2.5\cdot 10^{-3})$}  & \tiny{$0.62(0.01)$}   \\
\hline
HOG-Daublets  & \tiny{$0.12(3\cdot 10^{-3})$}  & \tiny{$0.28(3.8\cdot 10^{-3})$}  & \tiny{$0.15(3.7\cdot 10^{-3})$}  & \tiny{$0.65(9.5\cdot 10^{-3})$}  & \tiny{$0.28(2.9\cdot 10^{-3})$}  & \tiny{$0.16(2.9\cdot 10^{-3})$}  & \tiny{$0.64(7.5\cdot 10^{-3})$}   \\
\hline
HOG-Symmlets  & \tiny{$0.12(3.3\cdot 10^{-3})$}  & \tiny{$0.27(4.2\cdot 10^{-3})$} & \tiny{$0.15(5\cdot 10^{-3})$}  & \tiny{$0.64(8.6\cdot 10^{-3})$} & \tiny{$0.28(3.9\cdot 10^{-3})$}  & \tiny{$0.16(4.4\cdot 10^{-3})$}  & \tiny{$0.64(8.7\cdot 10^{-3})$}  \\
\hline
col.std &  \tiny{$0.077(1.93\cdot 10^{-3})$} & \tiny{$0.227(4.23\cdot 10^{-3})$} & \tiny{$0.131(5.74\cdot 10^{-3})$} & \tiny{$0.510( 0.012)$} & \tiny{$0.229(4.02\cdot 10^{-3})$} & \tiny{$0.135(5.37\cdot 10^{-3})$} &  \tiny{$ 0.509(0.014\cdot 10^{-3})$} \\
\hline
%P-splines & \tiny{$0.228(2.30\cdot 10^{-3})$} & \tiny{$0.306(6.79\cdot 10^{-3})$} & \tiny{$0.172(7.06\cdot 10^{-3})$} &   \tiny{$0.703(0.013)$} & \tiny{$0.319(4.83\cdot 10^{-3})$} & \tiny{$0.197(5.42\cdot 10^{-3})$} & \tiny{$0.680(0.011)$}  \\\hline
%P-splines:col.std & \tiny{$0.063(2.30\cdot 10^{-3})$} & \tiny{$0.237(3.85\cdot 10^{-3})$} & \tiny{$0.116(4.88\cdot 10^{-3})$} & \tiny{$0.596(0.012)$} & \tiny{$0.239(3.53\cdot 10^{-3})$} & \tiny{$0.119(4.40\cdot 10^{-3})$} &  \tiny{$0.596(0.011)$} \\\hline
col.std:HOG & \tiny{$0.051(1.99\cdot 10^{-3})$} & \tiny{$0.230(3.61\cdot 10^{-3})$} & \tiny{$0.131(5.76\cdot 10^{-3})$} & \tiny{$0.523(0.015)$} & \tiny{$0.232(3.29\cdot 10^{-3})$} & \tiny{$0.133(4.97\cdot 10^{-3})$}  & \tiny{$0.523( 0.013\cdot 10^{-3})$}  \\
\hline
\multicolumn{8}{|c|}{{\color{orange}Channel: $[f=64,~\psi=\pi]$}} \\
\hline
Feature vector & Entropy & MER & FPR & FNR  & probMER & probFPR & probFNR  \\ 
\hline
EDF      & \tiny{${\bf 1.01\cdot 10^{-8}}(2.19\cdot 10^{-8})$} &  \tiny{${\bf\approx 0}(0)$} &  \tiny{${\bf\approx 0}(0)$} & \tiny{${\bf\approx 0}(0)$}  &
          \tiny{${\bf 8.15\cdot 10^{-10}}(1.82\cdot 10^{-9})$} &  \tiny{${\bf 2.71\cdot 10^{-10}}(7.95\cdot 10^{-10})$} &   \tiny{${\bf 2.42\cdot 10^{-9}}(5.06\cdot 10^{-9})$} \\
          \hline
%edf(unscaled)      & \tiny{$1.78 \cdot 10^{-8}(4.56\cdot 10^{-8})$} & \tiny{$0(0)$} &  \tiny{$0(0)$} & \tiny{$0(0)$} & \tiny{$1.63\cdot 10^{-9}(4.30\cdot 10^{-9})$}  & \tiny{$2.02\cdot 10^{-11}(6.06\cdot 10^{-11})$} & \tiny{$6.40\cdot 10^{-9}(1.69\cdot 10^{-8})$}  \\   \hline
HOG &\tiny{$0.16(3.4\cdot 10^{-3})$}  & \tiny{$0.31(4.1\cdot 10^{-3})$}  & \tiny{$0.18(6.9\cdot 10^{-3})$} & \tiny{$0.69(9.1\cdot 10^{-3})$}  &\tiny{$0.32(3.7\cdot 10^{-3})$}  &\tiny{$0.19(6.5\cdot 10^{-3})$}  & \tiny{$0.69(7.3\cdot 10^{-3})$}   \\
\hline
LBP-HOG  & \tiny{$0.29(3.6\cdot 10^{-3})$}  & \tiny{$0.35(3.5\cdot 10^{-3})$}  & \tiny{$0.23(4.3\cdot 10^{-3})$}  & \tiny{$0.74(0.015)$}  & \tiny{$0.37(3\cdot 10^{-3})$}  & \tiny{$0.26(4\cdot 10^{-3})$}  & \tiny{$0.71(0.012)$}   \\
\hline
HOG:LBP-HOG &  \tiny{$0.16(2.5\cdot 10^{-3})$}   &  \tiny{$0.36(3.8\cdot 10^{-3})$}   &  \tiny{$0.23(3.5\cdot 10^{-3})$}   &  \tiny{$0.74(1\cdot 10^{-3})$}   &  \tiny{$0.36(4\cdot 10^{-3})$}   &  \tiny{$0.24(3.9\cdot 10^{-3})$}   &  \tiny{$0.73(0.01)$}    \\
\hline
HOG-Haar  & \tiny{$0.15(2.9\cdot 10^{-3})$}  & \tiny{$0.30(2.8\cdot 10^{-3})$}  & \tiny{$0.17(5\cdot 10^{-3})$} & \tiny{$0.68(0.01)$}  & \tiny{$0.31(2.3\cdot 10^{-3})$} & \tiny{$0.18(3.8\cdot 10^{-3})$}  & \tiny{$0.67(8.6\cdot 10^{-3})$}   \\
\hline
HOG-Daublets  & \tiny{$0.15(2.5\cdot 10^{-3})$} & \tiny{$0.32(7\cdot 10^{-3})$}  & \tiny{$0.19(9.5\cdot 10^{-3})$} & \tiny{$0.71(6.9\cdot 10^{-3})$}  & \tiny{$0.33(5.2\cdot 10^{-3})$}  & \tiny{$0.20(7\cdot 10^{-3})$} & \tiny{$0.70(6.4\cdot 10^{-3})$}   \\
\hline
HOG-Symmlets  & \tiny{$0.16(3\cdot 10^{-3})$}  & \tiny{$0.32(5.7\cdot 10^{-3})$} & \tiny{$0.19(6.3\cdot 10^{-3})$}  & \tiny{$0.71(8.5\cdot 10^{-3})$}  & \tiny{$0.33(4.2\cdot 10^{-3})$}  & \tiny{$0.21(5.6\cdot 10^{-3})$}  & \tiny{$0.70(7\cdot 10^{-3})$}   \\
\hline
col.std & \tiny{$0.132(2.40\cdot 10^{-3})$} & \tiny{$0.296(4.79\cdot 10^{-3})$} & \tiny{$0.186(4.68\cdot 10^{-3})$} & \tiny{$0.622(0.012)$} & \tiny{$0.300(3.82\cdot 10^{-3})$} & \tiny{$0.193(4.25\cdot 10^{-3})$} & \tiny{$0.615(8.12\cdot 10^{-3})$}  \\
\hline
%P-splines & \tiny{$0.228(2.73\cdot 10^{-3})$} & \tiny{$0.326(6.99\cdot 10^{-3})$} & \tiny{$0.206(7.44\cdot 10^{-3})$} & \tiny{$0.680(9.27\cdot 10^{-3})$} & \tiny{$0.338(4.65\cdot 10^{-3})$} & \tiny{$0.229(5.02\cdot 10^{-3})$} & \tiny{$0.662(7.91\cdot 10^{-3})$}  \\\hline
%P-splines:col.std & \tiny{$0.089(2.23\cdot 10^{-3})$} & \tiny{$0.299(4.84\cdot 10^{-3})$} & \tiny{$0.176(6.09\cdot 10^{-3})$} & \tiny{$0.663(0.012)$} & \tiny{$0.301(4.39\cdot 10^{-3})$} & \tiny{$0.180(5.46\cdot 10^{-3})$} & \tiny{$0.660(0.011)$} \\\hline
col.std:HOG &  \tiny{$0.101(1.87\cdot 10^{-3})$} & \tiny{$0.307(3.89\cdot 10^{-3})$} & \tiny{$0.205(4.95\cdot 10^{-3})$} & \tiny{$0.608(0.015\cdot 10^{-3})$} & \tiny{$0.310(3.51\cdot 10^{-3})$} & \tiny{$0.210(4.36\cdot 10^{-3})$}  & \tiny{$0.606(0.013)$}  \\
\hline
\end{tabular}
%\end{table}
\end{sidewaystable}

\subsection{3-class classification problem}

To investigate the defect detection capability of the proposed approach for finding the most common types of defects such as crater and dirt, we examined a 3-class classification problem. Among the alternative features considered in this paper, the feature vector based on HOG produced the best results when solving the binary classification task. Figures \ref{hOG_2classes}-\ref{fig:edf_3_class} (Appendix \ref{section:appendix}) show the columns means by class of the channel $[f=16,~\psi=\pi]$ for the HOG and EDF feature vectors for the binary and 3-class cases correspondingly.
So, the performance results of the HOG are shown here together with those of the EDF. The classification results of the probabilistic $k$-NN are presented in Table \ref{tab:knn-3-class}. The SVM classifier using HOG features failed to distinguish between three classes, at the same time SVM on the EDF performed as good as $k$-NN on EDF. So, the results on SVM performance are not included here. Table \ref{tab:knn-3-class} demonstrates that almost 100\% of defect detection and 0\% of false alarms are achieved when applying the proposed approach. Equivalently, nearly all patches with both crater and dirt were correctly classified, and no defect-free patch was predicted as defective. Contrary to the EDF, the HOG descriptors are capable of defect detection to a much lower degree, accounting for more than 10\% of false alarms and higher than 50\% of non-detections. The performance results of the HOG are worsening for the datasets of patches that correspond to channels with larger values of the frequency parameter, whereas the EDF performs equally good for all presented datasets. 

%\begin{table}[H]
%\scriptsize
\begin{sidewaystable}
\footnotesize
\caption{\label{tab:knn-3-class} Probabilistic $k$-NN results for 3-class classification problem. Performance results are means over ten runs. The corresponding standard errors are given in brackets.}
\centering
\begin{tabular}{|c||c|c|c|c|c|c|c|}
\hline
\multicolumn{8}{|c|}{{\color{orange}Channel: $[f=8,~\psi=\pi]$}} \\
\hline
Feature vector & Entropy & MER & FPR & FNR  & probMER & probFPR & probFNR  \\ 
\hline
EDF      & \tiny{${\bf 1.53\cdot 10^{-6}}(1.60\cdot 10^{-6})$} &  \tiny{${\bf\approx 0}(0)$} &  \tiny{${\bf\approx 0}(0)$}  &   \tiny{${\bf\approx 0}(0)$} & \tiny{${\bf 1.91\cdot 10^{-7}}(2.14\cdot 10^{-7})$}   & \tiny{${\bf 2.18\cdot 10^{-7}}(2.77\cdot 10^{-7})$} &  \tiny{${\bf 1.11\cdot 10^{-7}}(2.30\cdot 10^{-7})$} \\
\hline
HOG & \tiny{$0.078(3.2\cdot 10^{-3})$}  & \tiny{$0.24(5.2\cdot 10^{-3})$}  &\tiny{$0.11(5\cdot 10^{-3})$}  & \tiny{$0.54(0.011)$}  & \tiny{$0.24(5\cdot 10^{-3})$}  & \tiny{$0.12(5\cdot 10^{-3})$}  & \tiny{$0.61(0.012)$}   \\
\hline
\multicolumn{8}{|c|}{{\color{orange}Channel: $[f=16,~\psi=\pi]$}} \\
\hline

Feature vector & Entropy & MER & FPR & FNR  & probMER & probFPR & probFNR  \\
\hline
  EDF      & \tiny{${\bf 2.17\cdot 10^{-5} }(3.73\cdot 10^{-5})$} &  \tiny{${\bf\approx 0}(0)$} &  \tiny{${\bf\approx 0}(0)$}  &   \tiny{${\bf\approx 0}(0)$} & \tiny{${\bf9.57\cdot 10^{-6}}(2.49\cdot 10^{-5})$}   & \tiny{${\bf1.28\cdot 10^{-6}}(2.64\cdot 10^{-6})$} &  \tiny{${\bf3.41\cdot 10^{-5}}(9.95\cdot 10^{-5})$}  \\
\hline
HOG &\tiny{$0.089(1.2\cdot 10^{-3})$}  & \tiny{$0.25(3.3\cdot 10^{-3})$}  & \tiny{$0.12(4.1\cdot 10^{-3})$} & \tiny{$0.58(7.9\cdot 10^{-3})$} &\tiny{$0.25(2.5\cdot 10^{-3})$}  &\tiny{$0.13(2.9\cdot 10^{-3})$} & \tiny{$0.64(7.9\cdot 10^{-3})$}   \\
\hline
\multicolumn{8}{|c|}{{\color{orange}Channel: $[f=32,~\psi=\pi]$}} \\
\hline
Feature vector & Entropy & MER & FPR & FNR  & probMER & probFPR & probFNR  \\
\hline
  EDF      & \tiny{${\bf 1.28\cdot 10^{-5}}(1.60\cdot 10^{-5})$} &  \tiny{${\bf\approx 0}(0)$} &  \tiny{${\bf\approx 0}(0)$}  &   \tiny{${\bf\approx 0}(0)$} & \tiny{${\bf 2.72\cdot 10^{-6}}(4.33\cdot 10^{-6})$}   & \tiny{${\bf 2.98\cdot 10^{-6}}(5.99\cdot 10^{-6})$} &  \tiny{${\bf 1.96\cdot 10^{-6}}(3.93\cdot 10^{-6})$}  \\
\hline
HOG &\tiny{$0.12(3.2\cdot 10^{-3})$}  &\tiny{$0.28(4.5\cdot 10^{-3})$}  &\tiny{$0.15(5.1\cdot 10^{-3})$}  &\tiny{$0.65(0.012)$}  & \tiny{$0.29(3.9\cdot 10^{-3})$}  & \tiny{$0.15(4.7\cdot 10^{-3})$}  & \tiny{$0.69(7.6\cdot 10^{-3})$}  \\
\hline
\multicolumn{8}{|c|}{{\color{orange}Channel: $[f=64,~\psi=\pi]$}} \\
\hline
Feature vector & Entropy & MER & FPR & FNR  & probMER & probFPR & probFNR  \\ 
\hline
EDF      &  \tiny{${\bf 2.50\cdot 10^{-8}}(3.10\cdot 10^{-8})$} &  \tiny{${\bf \approx 0}(0)$} &  \tiny{${\bf\approx 0}(0)$}  &   \tiny{${\bf\approx 0}(0)$} & \tiny{${\bf 2.04\cdot 10^{-9}}(2.62\cdot 10^{-9})$}   & \tiny{${\bf 6.18\cdot 10^{-10}}(1.87\cdot 10^{-9})$} &  \tiny{${\bf 6.26\cdot 10^{-9}}(9.11\cdot 10^{-9})$}  \\
\hline
HOG & \tiny{$0.17(2.3\cdot 10^{-3})$}  & \tiny{$0.32(4.8\cdot 10^{-3})$}  & \tiny{$0.18(5.1\cdot 10^{-3})$} & \tiny{$0.67(0.01)$}  &\tiny{$0.33(4.2\cdot 10^{-3})$}  & \tiny{$0.19(4.7\cdot 10^{-3})$}  & \tiny{$0.73(7.5\cdot 10^{-3})$} \\
\hline
\end{tabular}
%\end{table}
\end{sidewaystable}

\section{Conclusions}\label{section:conclusions}

In this paper we have introduced an efficient statistical learning approach for defect detection on specular surface using deflectometry-based images. Our approach is build on the smoothness degree of the splines fitted to pixel values of the images. Classification is achieved using a probabilistic $k$-NN classifier that allows examining classification quality judgment. 

The proposed statistical learning approach has been applied to the experimental industrial setup installed at Volvo GTO Cab plant in Ume{\aa}, Sweden. 
Our method appears to be successful in addressing the defect detection and classification problem on specular cab surfaces. It correctly detects defects and is capable of distinguishing crater and dirt.
In addition, the probability based performance evaluation metrics have been proposed as alternatives to the conventional metrics. The usage of the probability based metrics allows for uncertainty estimation of the predictive performance of a classifier.
The experimental results showed that our approach outperforms conventional existing approaches in terms of misclassification error rate, false positive and false negative rates and their probabilistic analogues, and not least the robustness with respect to various image channels and patch sizes.     

The further research will focus on developing a real-time pilot system for defect detection and classification with an efficient and automated statistical process quality control system. %The self-driven system is currently running at Volvo GTO cab plant in Ume{\aa} and the preliminary results look very promising. 

%\todo{Here's a comment in the margin!}, as shown in the example on the right. You can also add inline comments:

%\todo[inline, color=green!40]{This is an inline comment.}

\section*{Acknowledgements}
This work is a part of FIQA project, supported by the Strategic Vehicle Program Research and Innovation (FFI) at VINNOVA - Sweden's Innovation Agency (Reg. No. 2015-03706) and Volvo Group Trucks Operations.       

\bibliography{main}

\appendix
\section{Additional figures}\label{section:appendix}

\begin{figure}[H]
\begin{centering}
\includegraphics[scale=0.35]{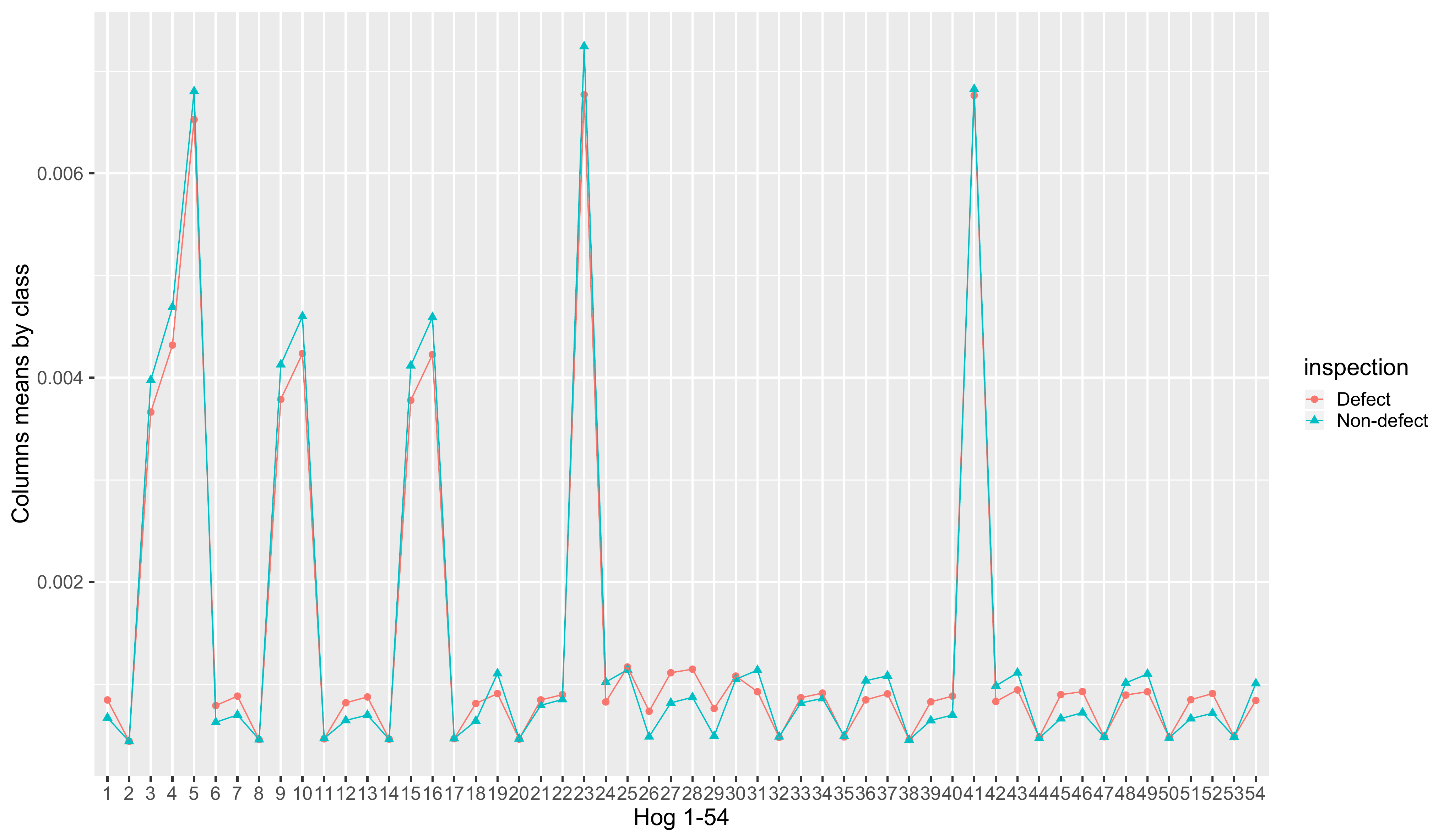}
\par\end{centering}
\caption{\label{hOG_2classes} HOG 2-class feature vector (columns means by class, channel $[f=16,~\psi=\pi]$).}
\end{figure}

\begin{figure}[H]
\centering
\includegraphics[width=.95\textwidth,height=0.4\textheight]{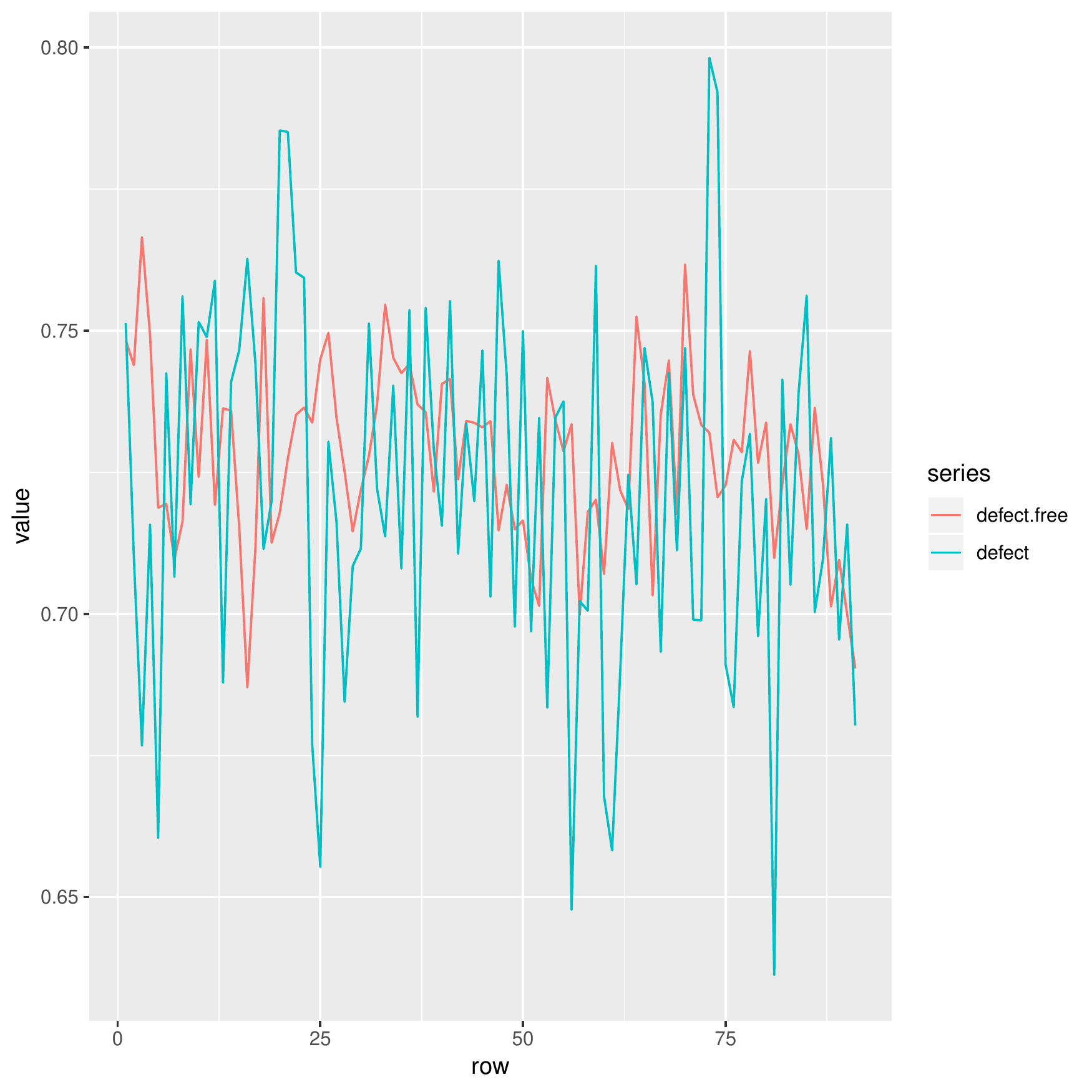}
\caption{\label{fig:edf_2_class}EDF 2-class feature vector (columns means by class, channel $[f=16,~\psi=\pi]$).}
\end{figure}

\begin{figure}[H]
\begin{centering}
\includegraphics[scale=0.35]{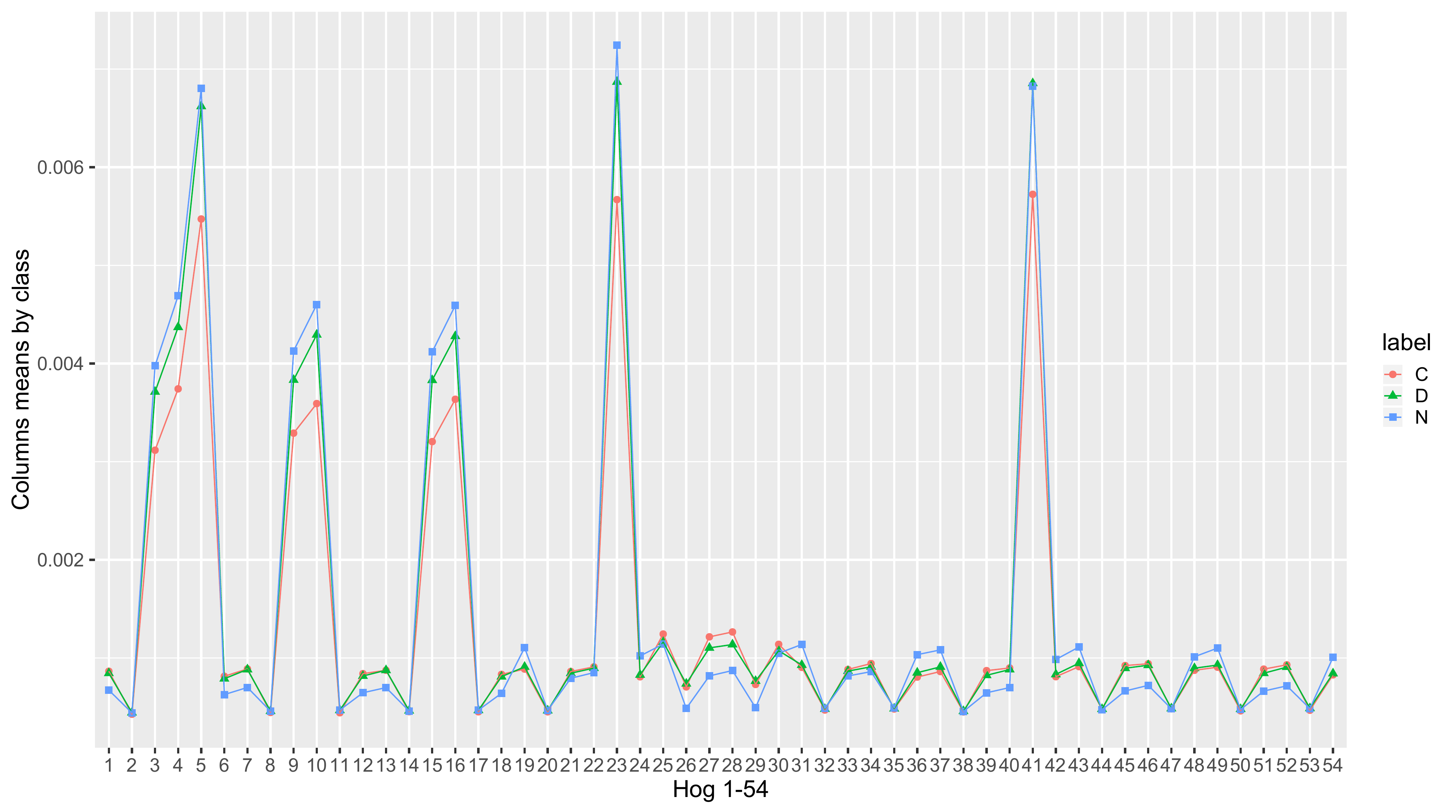}
\par\end{centering}
\caption{\label{hOG_3classes} HOG 3-class feature vector (columns means by class, channel $[f=16,~\psi=\pi]$).}
\end{figure}

\begin{figure}[H]
\centering
\includegraphics[width=.95\textwidth,height=0.4\textheight]{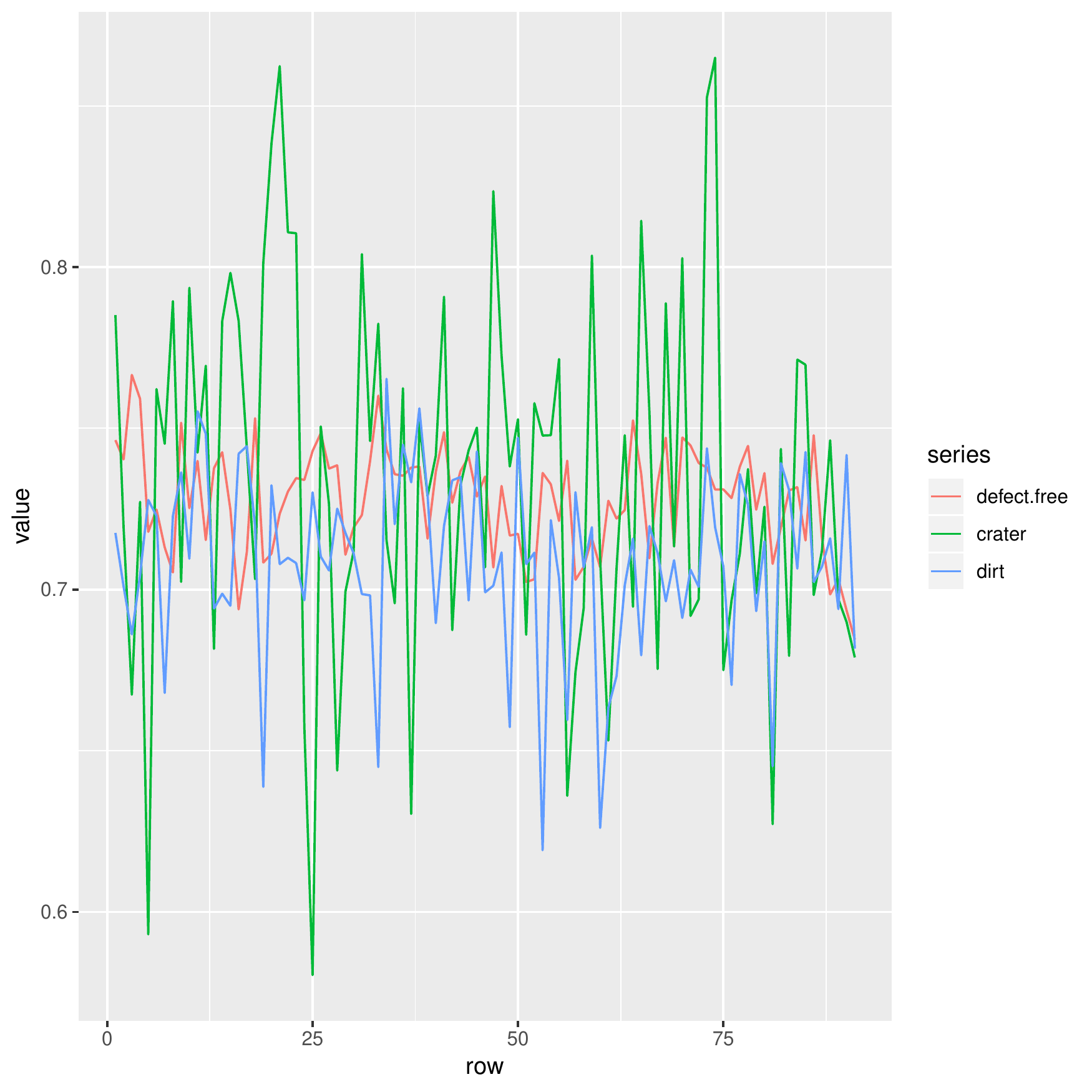}
\caption{\label{fig:edf_3_class}EDF 3-class feature vector (columns means by class, channel $[f=16,~\psi=\pi]$).}
\end{figure}

\end{document}